\documentclass[twoside]{article}

\usepackage{microtype}

\usepackage{graphicx}
\usepackage{subfig}
\usepackage{booktabs} 

\usepackage[colorlinks=true, linkcolor=black, citecolor=black, urlcolor=black]{hyperref}

\usepackage[ruled,vlined,linesnumbered,algo2e]{algorithm2e}

\usepackage{mdframed}
\usepackage{xcolor}
\definecolor{lightgray}{gray}{0.95}
\newmdenv[
  backgroundcolor=lightgray,
  linecolor=gray,
  linewidth=0.0pt,
  skipabove=0.0pt,
  skipbelow=0.5pt,
  innerleftmargin=6pt,
  innerrightmargin=6pt,
  innertopmargin=-0.5em,
  innerbottommargin=4pt
]{graybox}

\usepackage[accepted]{aistats2026}

\usepackage{amsmath}
\usepackage{amssymb}
\usepackage{mathtools}
\usepackage{amsthm}
\usepackage[ruled,vlined]{algorithm2e}
\usepackage{algorithm}
\usepackage{algorithmic}

\usepackage{color}
\usepackage{booktabs}

\usepackage{blindtext}
\usepackage{enumerate}
\usepackage{graphicx}
\usepackage{amsfonts, amsmath, bm, amssymb}
\usepackage{dsfont}
\usepackage{wrapfig}
\usepackage{enumitem}
\usepackage{booktabs}
\usepackage{multirow}
\usepackage{xfrac}
\usepackage{makecell}
\usepackage{overpic}
\newcommand\numberthis{\addtocounter{equation}{1}\tag{\theequation}}
\usepackage{mwe}
\usepackage{listings}
\definecolor{darkgreen}{RGB}{0, 128, 27}
\usepackage{placeins}
\usepackage{thm-restate}

\def\argmin{\textrm{argmin}}
\def\proj{\textrm{proj}}

\def\conv{\hspace{-0.1em}*\hspace{-0.1em}}

\newcommand{\zero}{\mathbf{0}}

\newcommand{\diff}{\,\mathrm{d}}

\newcommand{\inner}[2]{\left\langle #1, #2 \right\rangle}

\newcommand{\KL}[2]{D_{\scriptscriptstyle\mathrm{KL}}({#1}\hspace{0.1em} \|\hspace{0.1em}  {#2})}

\usepackage{xspace}
\makeatletter
\DeclareRobustCommand\onedot{\futurelet\@let@token\@onedot}
\def\@onedot{\ifx\@let@token.\else.\null\fi\xspace}

\makeatother

\newcommand{\Bc}{\mathcal{B}}

\newcommand{\Dc}{\mathcal{D}}
\newcommand{\Ec}{\mathcal{E}}
\newcommand{\Fc}{\mathcal{F}}

\newcommand{\Kc}{\mathcal{K}}
\newcommand{\Lc}{\mathcal{L}}
\newcommand{\Mc}{\mathcal{M}}
\newcommand{\Nc}{\mathcal{N}}

\newcommand{\Tc}{\mathcal{T}}

\newcommand{\Yc}{\mathcal{Y}}

\newcommand{\Eb}{\mathbb{E}}

\newcommand{\Nb}{\mathbb{N}}

\newcommand{\Rb}{\mathbb{R}}

\newcommand{\bv}{\mathbf{b}}

\newcommand{\fv}{\mathbf{f}}

\newcommand{\sv}{\mathbf{s}}

\newcommand{\uv}{\mathbf{u}}

\newcommand{\wv}{\mathbf{w}}
\newcommand{\xv}{\mathbf{x}}
\newcommand{\yv}{\mathbf{y}}

\newcommand{\Iv}{\mathbf{I}}

\newcommand{\epsilonv   }{\boldsymbol \epsilon   }

\newcommand{\thetav     }{\boldsymbol \theta     }

\newcommand{\nuv        }{\boldsymbol \nu        }

\newcommand{{\phiv}     }{\boldsymbol \phi       }

\ifx\BlackBox\undefined
\newcommand{\BlackBox}{\rule{1.5ex}{1.5ex}}  
\fi
\ifx\QED\undefined
\def\QED{~\rule[-1pt]{5pt}{5pt}\par\medskip}
\fi
\ifx\proof\undefined
\newenvironment{proof}{\par\noindent{\em Proof:\ }}{\hfill\BlackBox\\}
\fi
\ifx\theorem\undefined
\newtheorem{theorem}{Theorem}
\fi
\ifx\example\undefined
\newtheorem{example}{Example}
\fi
\ifx\property\undefined
\newtheorem{property}{Property}
\fi
\ifx\lemma\undefined
\newtheorem{lemma}{Lemma}
\fi
\ifx\proposition\undefined
\newtheorem{proposition}{Proposition}
\fi
\ifx\fact\undefined
\newtheorem{fact}{Fact}
\fi
\ifx\remark\undefined
\newtheorem{remark}{Remark}
\fi
\ifx\corollary\undefined
\newtheorem{corollary}{Corollary}
\fi
\ifx\definition\undefined
\newtheorem{definition}{Definition}
\fi
\ifx\conjecture\undefined
\newtheorem{conjecture}{Conjecture}
\fi
\ifx\axiom\undefined
\newtheorem{axiom}[theorem]{Axiom}
\fi
\ifx\claim\undefined
\newtheorem{claim}[theorem]{Claim}
\fi
\ifx\assumption\undefined
\newtheorem{assumption}{Assumption}
\fi
\ifx\question\undefined
\newtheorem{question}{Question}
\fi
\ifx\problem\undefined
\newtheorem{problem}{Problem}
\fi

\usepackage[capitalize,noabbrev]{cleveref}
\usepackage{textcomp} 
\crefformat{prop}{Prop~#2#1#3}
\crefformat{figure}{Fig~#2#1#3}
\crefformat{lem}{Lem~#2#1#3}
\crefformat{lemma}{Lem~#2#1#3}
\crefformat{section}{Sec~#2#1#3}
\crefformat{equation}{Eq~(#2#1#3)}
\crefformat{algorithm}{Alg~#2#1#3}
\crefformat{fact}{Fact~#1}
\crefformat{chapter}{Chapter~#2#1#3}
\crefalias{lem}{Lemma}
\crefformat{theorem}{Thm~#1}
\crefformat{corollary}{Cor~#2#1#3}
\crefformat{proposition}{Prop~#2#1#3}
\Crefname{appendix}{Appx}{Appx}

\usepackage[textsize=tiny]{todonotes}

\usepackage[round]{natbib}

\begin{document}

\runningtitle{SFBD Flow: Training Diffusion Models with Noisy Samples}

\twocolumn[
\aistatstitle{SFBD Flow: A Continuous-Optimization Framework \\for Training Diffusion Models with Noisy Samples}

\aistatsauthor{ Haoye Lu \And Darren Lo  \And  Yaoliang Yu }

\aistatsaddress{ University of Waterloo \\ Vector Institute \And University of Waterloo \And University of Waterloo \\ Vector Institute} 

]

\begin{abstract}
Diffusion models achieve strong generative performance but often rely on large datasets that may include sensitive content. This challenge is compounded by the models’ tendency to memorize training data, raising privacy concerns. SFBD (Lu et al., 2025) addresses this by training on corrupted data and using limited clean samples to capture local structure and improve convergence. However, its iterative denoising and fine-tuning loop requires manual coordination, making it burdensome to implement. We reinterpret SFBD as an alternating projection algorithm and introduce a continuous variant, SFBD flow, that removes the need for alternating steps. We further show its connection to consistency constraint-based methods, and demonstrate that its practical instantiation, Online SFBD, consistently outperforms strong baselines across benchmarks.
\end{abstract}

\section{INTRODUCTION}
\label{sec:intro}

Diffusion-based generative models \citep{DicksteinWMG2015, HoJA2020, SongME2021, SongDCKKEP2021, SongDCS2023} have attracted growing interest and are now regarded as one of the most powerful frameworks for modelling high-dimensional distributions. They have enabled remarkable progress across various domains \citep{CroitoruHIS2023}, including image \citep{HoJA2020, SongME2021, SongDCKKEP2021, RombachBLEO2022}, audio \citep{kong2021diffwave, YangYWWWZY2023}, and video generation \citep{HoSGCNF2022}.

Diffusion models can be efficiently trained using the conditional score-matching loss, making them relatively easy to scale. This scalability enables the training of very large models on web-scale datasets -- a crucial factor in achieving high performance. This approach has driven recent breakthroughs in image generation, exemplified by models such as Stable Diffusion (-XL) \citep{RombachBLEO2022, PodellELBDMPR2023} and DALL-E  \citep{BGJBWLOZLG2023}. However, this success comes with challenges: large-scale datasets often include copyrighted material, and diffusion models are more prone than earlier generative methods like GANs \citep{GoodfellowPMXWOCB2014, GoodfellowPMXWOCB2020} to memorizing training data, potentially reproducing entire samples \citep{CarliniHNJSTBIDW2023, SomepalliSGGG2023}.


A recently proposed strategy to address memorization and copyright concerns involves training or fine-tuning diffusion models on corrupted data \citep{DarasSDGDK2023, SomepalliSGGG2023, DarasA2023, DarasDD2024}. In this setting, the model never has direct access to the original data. Instead, each sample is transformed via a known, non-invertible corruption process, such as pixel-wise additive Gaussian noise in image datasets, ensuring that the original content cannot be reconstructed or memorized at the individual sample level. Remarkably, under mild conditions, such corruptions - although irreversible at the sample level -- can induce a bijection between the original and corrupted data distributions \citep{BoraPD2018}. Specifically, the corrupted data distribution has a density equal to the convolution of the true data density with the corruption noise distribution \citep{Meister2009, LuWY2025}. As a result, it is theoretically possible to recover the original data distribution by first estimating the corrupted (noisy) density from samples, and then performing density deconvolution to approximate the underlying true data density.

We refer to this task -- recovering the true data distribution from noisy observations -- as the \textit{deconvolution problem}. Motivated by this formulation, several works \citep{DarasDD2024, DarasCD25, LuWY2025} have shown that diffusion models can effectively address the deconvolution problem either by applying iterative denoising and fine-tuning, as in SFBD \citep{LuWY2025}, or by enforcing consistency constraints (CCs) during training \citep{DarasDDD2023}. Specifically, when paired with a small set of copyright-free clean data, both SFBD and CC-based methods have been shown to guide diffusion models toward generating high-quality images. However, SFBD relies on costly iterative denoising and fine-tuning, while CC-based methods require solving backward stochastic differential equations (SDEs) at each training step, making both approaches computationally expensive in different ways.

In this paper, we eliminate the iterative denoising and fine-tuning required by SFBD by introducing a continuous variant, \emph{SFBD flow}. We reinterpret SFBD as alternating projections between two sets of stochastic processes, framing it as a stochastic process optimization problem. Inspired by Schrödinger bridge flow~\citep{BortoliKMD2024}, this perspective yields a generalized family of diffusion-based deconvolution methods, $\gamma$-SFBD for $\gamma \in (0,1]$, that guide the model toward the clean data distribution. Standard SFBD is recovered when $\gamma=1$, while letting $\gamma \to 0$ turns the discrete sequence into a continuous evolution, giving rise to the SFBD flow. We further show that SFBD flow corresponds to steepest descent in function space, with $\gamma$-SFBD as its discrete approximation using step size $\gamma$. This interpretation motivates Online SFBD, a practical diffusion-based deconvolution method that avoids repeated fine-tuning (\cref{sec:online_SFBD}). By adaptively adjusting $\gamma$, Online SFBD accommodates approximation errors from imperfect training, achieving faster and more stable convergence. We also reveal a close connection to CC-based methods, providing a unified perspective. Empirical results validate our analysis, with Online SFBD consistently outperforming strong baselines across benchmarks. The implementation is available at \href{https://github.com/watml/SFBD-flow}{https://github.com/watml/SFBD-flow}.

\section{PRELIMINARIES}
\label{sec:prel}
In this section, we review diffusion models, the deconvolution problem, and two typical methods for training diffusion models on data corrupted by Gaussian noise.

\textbf{Diffusion Models.} Diffusion models generate data by progressively adding Gaussian noise to input samples and then learning to reverse this process through a sequence of denoising steps. Formally, given an initial data distribution $p_0$ over $\Rb^d$, the forward process is defined by the SDE
\begin{align}
	\diff \xv_t = \diff \wv_t, ~~\textrm{$\xv_0 \sim p_0$,~ $t\in [0, T]$,} \label{eq:fwd_diff}
\end{align}
where $T$ is a fixed positive constant.  $\{\wv_t\}_{t\in [0,T]}$ is the standard Brownian motion.\footnote{Despite its simplicity, nearly all existing diffusion models adopt a forward process equivalent to \cref{eq:fwd_diff}, as shown in \cref{sec:eqv_fwd_proc}. }

 \cref{eq:fwd_diff} induces a transition kernel $p_{t|s}(\xv_t | \xv_s)$ for $0 \leq s \leq t \leq T$.  For $s = 0$, 
\begin{align}
    p_{t|0}(\xv_t | \xv_0) = \Nc(\xv_0, t \, \Iv), ~~\textrm{~ $t\in [0, T]$.} \label{def:diff_trans_kernel}
\end{align}  
When $T$ is large, the terminal state $\xv_T$ closely approximates a sample from the isotropic Gaussian distribution $\Nc(\zero, T\, \Iv)$. Let $ p_t(\xv_t) = \int p_{t|0}(\xv_t \vert \xv_0) \, p_0(\xv_0) \, \mathrm{d} \xv_0 $ denote the marginal distribution of $\xv_t$, where $p_T \approx \Nc(\zero, T\, \Iv)$. \citet{anderson1982} showed that the time-reversed dynamics of the forward SDE can be expressed as the backward SDE:
\begin{align}
	\diff \xv_t = - \sv_t(\xv_t) \diff t + \diff \bar\wv_t, ~\xv_T \sim p_T \label{eq:anderson_bwd},
\end{align}
where $\bar{\wv}_t$ is standard Brownian motion in reverse time and $\sv_t = \sv^*_t := \nabla \log p_t$ is the score function. Crucially, this reverse SDE induces transition kernels that match the posterior of the forward process: $p_{s|t}(\xv_s | \xv_t) = \tfrac{p_{t|s}(\xv_t | \xv_s)  p_s(\xv_s)}{p_t(\xv_t)}$ for $s \leq t$ in $[0,T]$. It is known that $\sv^*_t(\xv_t) =  \tfrac{1}{t}(\Eb_{p_{0|t}}[\xv_0 \vert \xv_t] - \xv_t)$, where the conditional expectation $\Eb_{p_{0|t}}[\xv_0 \mid \xv_t]$ is typically approximated in practice by a neural network-denoiser $D_{\phiv}(\xv_t)$ \citep{KarrasAAL22}, trained by minimizing
\begin{align}
	\Lc_d(\phiv) = {\Eb}_{t \sim \Tc}\, {\Eb}_{p_0} \, {\Eb}_{p_{t|0}} \left[w(t) \|D_{\phiv}(\xv_t, t) - \xv_0 \|^2 \right], \label{eq:denoiser_loss} 
\end{align}
where $w(t)$ is a time-dependent weighting function and $\Tc$ denotes a sampling distribution over $[0, T]$. With a well-trained denoiser $D_{\phiv}$, $\sv^*_t$ can be approximated by
\begin{align}
	\sv^{\phiv}_t(\xv_t)  := \tfrac{1}{t} \big(D_{\phiv}(\xv_t, t) - \xv_t \big). \label{eq:score_in_denoise}
\end{align}
Substituting this estimate into \cref{eq:anderson_bwd}, one can simulate the reverse-time SDE starting from $\tilde{\xv}_T \sim \Nc(\zero,  T\, \Iv)$, yielding a sample $\tilde{\xv}_0$ that serves as an approximation sampled from~$p_0$.

\textbf{Deconvolution Problem.} We follow the setup of \citet{DarasDD2024, LuWY2025}, where corrupted samples $\Yc = \{\yv^{(i)}\}_{i=1}^{n}$ are generated as
\(
    \yv^{(i)} = \xv^{(i)} + \epsilonv^{(i)}, \label{eq:gen_conv_samples}
\)
with $\xv^{(i)} \sim p_{\rm data}$ and $\epsilonv^{(i)} \sim h = \Nc(\zero, \tau\Iv)$ drawn independently, where $\tau \in (0, T)$ is known and fixed. The resulting samples $\yv^{(i)}$ follow a distribution with density $p^*_\tau = p_{\rm data} \conv h$, where $\conv$ denotes the convolution operator \citep{LuWY2025}. In addition, we assume access to a small set of clean samples $\Dc_\text{clean} = \{\xv^{(i)}\}_{i=1}^M$ with $\xv^{(i)} \sim p_{\rm data}$.

While deconvolution theory \citep{Meister2009, LuWY2025} and related empirical results in the context of GANs \citep{BoraPD2018} have demonstrated the theoretical and practical feasibility of learning the true data distribution from noisy samples, a key challenge remains: how to effectively train a diffusion model on corrupted data to generate clean samples.

\textbf{Consistency Constraint-based Method}. \citet{DarasDD2024} first addressed this problem using CCs~\citep{DarasDDD2023}. With noisy samples $\xv_\tau \sim p^*_\tau$, they trained a network $\sv^{\phiv}_t$ to approximate score $\sv^*_t$ for $t > \tau$ via a modified loss called ambient score matching (ASM). Specifically, $\sv^{\phiv}_t$ is implemented through \cref{eq:score_in_denoise}, where $D_{\phiv}(\xv_t, t)$ approximates $\Eb_{p_{0|t}}[\xv_0 \mid \xv_t]$. For $t \leq \tau$, score matching is inapplicable, and instead $D_{\phiv}(\xv_t, t)$ is trained to satisfy the CCs:
\begin{align*}
	\Eb_{p_{0|s}}[\xv_0|\xv_s] =  {\Eb}_{p_{r\vert s}} \big[\Eb_{p_{0|r}}[\xv_0|\xv_r] \big], \text{ for $0 \leq r \leq s \leq T$}
\end{align*} 
by jointly minimizing the \textit{consistency loss}:
\begin{align}
	\hspace{-0.5em}\Lc_\text{con}({\phiv}, r, s) \hspace{-0.125em}=\hspace{-0.125em} \Eb_{p_s} \big\|D_{\phiv}(\xv_s, s) - \Eb_{p_{r|s} } [D_{\phiv}(\xv_r, r)]  \big\|^2 \label{eq:consistency_loss}
\end{align}
where $r$ and $s$ are sampled from predefined distributions. Sampling from $p_{r|s}$ is performed by solving \cref{eq:anderson_bwd} backward from $\xv_s$, using the network-estimated drift $\sv_t^{\phiv}$ from \cref{eq:score_in_denoise}. To sample from $p_s$, one first draws $\xv_\tau$ for $\tau > s$, then samples $\xv_s \sim p_{s|\tau}$ analogously. If $D_{\phiv}$ minimizes the consistency loss for all $r, s$ and satisfies $\sv_t^{\phiv} = \sv_t^*$ for $t > \tau$, then $\sv_t^{\phiv}$ exactly recovers $\sv_t^*$ for all $t \in [0, T]$, allowing $p_0 = p_\text{data}$ to be sampled via \cref{eq:anderson_bwd} \citep{DarasDD2024}.

However, both \citeauthor{DarasCD25} and \citeauthor{LuWY2025} showed that using CCs alone is insufficient to recover the drift below $\tau$ due to poor sample complexity \citep{LuWY2025, DarasCD25}. \citeauthor{DarasCD25} addressed this by adding the denoising loss \cref{eq:denoiser_loss} on $\Dc_\text{clean}$, achieving strong empirical results.

\textbf{Stochastic Forward-backward Deconvolution.} Instead of relying on CCs to recover the distribution for $t \leq \tau$, \citet{LuWY2025} proposed an iterative scheme, SFBD, that alternates between finetuning and denoising steps. Given a sample set $\Ec$, let $p_\Ec$ denote the empirical distribution induced by $\Ec$. Starting from a pretrained model $D_{\phiv_0}$ trained on $\Dc_\text{clean}$, the algorithm proceeds as follows for $k = 1, 2, \ldots, K$:

(Denoise) $\Ec_k \leftarrow \{ \yv_0^{(i)}:$ solve \cref{eq:anderson_bwd} from $t = \tau$ to $0$ with  $\sv_t(\xv_t) = \tfrac{D_{\phiv_k}(\xv_t, t) - \xv_t}{t}, \xv_\tau = \yv_\tau^{(i)} \in \Ec_\text{noisy} \}$.

(Finetune) Update $D_{\phiv_k}$ to obtain $D_{\phiv_{k+1}}$ by miminizing \cref{eq:denoiser_loss} with $p_0 = p_{\Ec_k}$.

\citet{LuWY2025} showed that as $K \to \infty$, $p_{\Ec_K}$ converges to the true distribution $p_\text{data}$. While SFBD outperforms DDIM~\citep{SongME2021} trained solely on clean data (e.g., on CelebA~\citep{LiuLWT2015}), its iterative nature makes implementation challenging. 

\section{SFBD AS ALTERNATIVE PROJECTIONS}
\label{sec:SFBD_alt_proj}
In this section, we show that SFBD can be interpreted as an alternating projection algorithm. 

\textbf{Notation.} Let $\Mc$ be the set of path measures on $t \in [0,\tau]$ induced by the backward process~\cref{eq:anderson_bwd}, with arbitrary drift $\sv: [0,\tau]\times \Rb^d \to \Rb^d$ and fixed terminal distribution $p_\tau^\ast$ at $t=\tau$. We write $M(\sv)\in\Mc$ for the path measure corresponding to drift $\sv$. Likewise, let $\Dc$ be the set of path measures on $t \in [0,\tau]$ induced by the forward process~\cref{eq:fwd_diff} with arbitrary initial distribution $p_0$, and denote by $D(q)\in\Dc$ the measure induced by $p_0=q$. In particular, let $P^\ast = D(p_{\text{data}})$ and $\sv_t^\ast = \nabla \log p_t$ be the score function of the forward diffusion process initialized at $p_0=p_{\text{data}}$. Then $P^\ast = D(p_{\text{data}})=M(\sv^\ast)$, showing that $P^\ast$ lies in $\Mc \cap \Dc$, and is in fact the unique element therein.

\textbf{Alternative Projections.} SFBD then can be formulated as an algorithm alternating between two projections: the Markov projection (M-Proj) and the diffusion projection (D-Proj):
\begin{align}
\hspace{-1.7em}\text{(M-Proj)}\,
& M^k = \underset{\Mc}{\proj}\, P^k 
\hspace{-0.2em}:=\hspace{-0.2em} \underset{M \in \Mc}{\argmin}\, \KL{P^k}{M} 
\label{def:markov_proj}\hspace{-0.5em}\\[0.5em]
\hspace{-1.7em}\text{(D-Proj)}\, 
& P^{k+1} = \underset{\Dc}{\proj}\, M^k 
\hspace{-0.2em}:=\hspace{-0.2em} \underset{P \in \Dc}{\argmin}\, \KL{M^k}{P}
\label{def:diff_proj} \hspace{-0.9em}
\end{align}
for $k = 0, 1, 2, \ldots, K$, with initial path measure $P^0 = \Dc(p_{\Ec_\text{clean}})$. Since each $M \in \Mc$ is fully determined by a backward drift $\sv$, we denote the drift of $M^k$ by $\sv^k$, i.e., $M^k = M(\sv^k)$. Thus, the M-Proj can be equivalently written as $\argmin_\sv \KL{P^k}{M(\sv)}$.  

The M-Proj corresponds to the finetuning step in SFBD. To see this, by \cref{appx:lem:kl_two_path} in \cref{appx:aux_results},
\begin{align*}
	\KL{P^k}{M^k} = & {\KL{p^k_\tau}{p^*_\tau}} \\
	& \hspace{-1.4em}+  {\Eb}_{P^k}\big[\tfrac{1}{2\, } \textstyle  \int_0^\tau \| \nabla \log p^k_t(\xv_t)  - \sv_t^k(\xv_t) \|^2 \diff t \big]
\end{align*}
where $p_t^k$ denotes the marginal density of $P^k$ at time $t$. Since the first term is independent of $M^k$, minimizing the KL reduces to setting $\sv_t^k(\xv_t) = \nabla \log p_t^k(\xv_t)$, i.e., performing score matching. This corresponds to the fine-tuning step that minimizes \cref{eq:denoiser_loss} with $p_0 = p_0^k$~\citep{KarrasAAL22}.

\begin{figure}
	\includegraphics[width=\columnwidth]{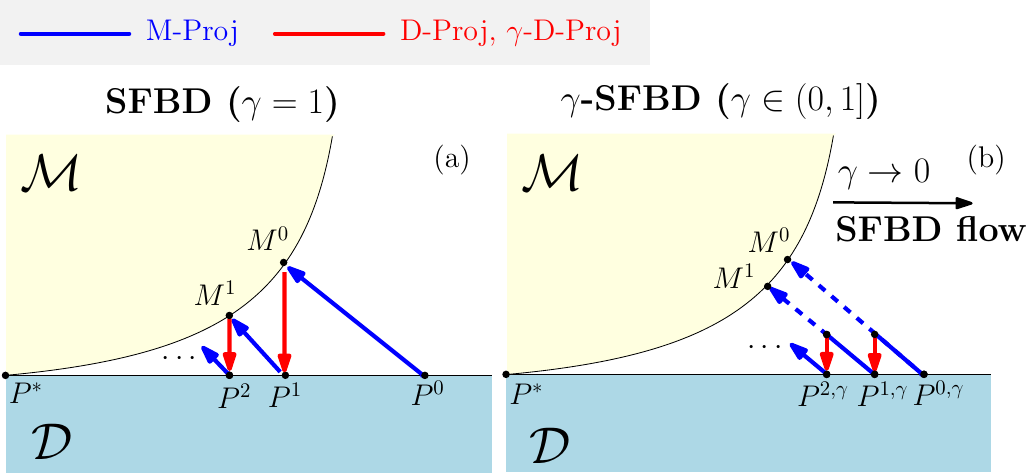}
	\caption{SFBD uses alternative projection to guide the stochastical process sequences $P^k$ and $M^k$ converge to the optimal $P^*$. When $\gamma \rightarrow 0$, the changes of $P^k$ and $M^k$ become smooth and we obtain $\gamma$-SFBD.}
	\label{fig:sfgd_proj}
\end{figure}

Likewise, D-Proj corresponds to the denoising step. By the disintegration theorem~\citep{VargasTLL2021},
\begin{equation}
\hspace{-0.25em}\KL{ M^k }{P}\hspace{-0.15em}=\hspace{-0.15em}\KL{m_0^k}{p_0}\hspace{-0.05em}+\hspace{-0.05em} \Eb_{M^k}\hspace{-0.3em}\left[\log\hspace{-0.1em}\tfrac{\diff M^k(\cdot \mid \xv_0)}{\diff P(\cdot \mid \xv_0)} \right] \label{eq:disintegration_thm_D_Proj}
\end{equation}
where $m_0^k$ is the marginal of $M^k$ at $t = 0$. Since $P \in \Dc$ is determined by the forward SDE in \cref{eq:fwd_diff}, its conditional path measure given $\xv_0$ is fixed, making the second term constant. Therefore, minimizing the KL divergence reduces to matching the marginals, i.e., $p_0 = m_0^k$ and thus $P^{k+1} = D(m_0^k)$. In other words, D-Proj sets $p_0$ to the distribution of the denoised samples in the denoising step. \cref{fig:sfgd_proj}a illustrates how SFBD applies the alternative projection algorithm to find $P^*$. 

\textbf{Evolution of $\sv^k$.} In practice, the only component in SFBD requiring estimation is $\sv^k = \nabla \log p_t^k(\xv_t)$, approximated by a neural network $\hat\sv^k$. As $k \to \infty$, $\sv_t^k$ converges to the true score function $\sv_t^* $ associated with the forward diffusion process~\cref{eq:fwd_diff} initialized with $p_0 = p_\text{data}$~\citep{LuWY2025}.

As the updates in Eqs~\eqref{def:markov_proj} and \eqref{def:diff_proj} can be compactly written as 
\(
    M^{k+1} = \argmin_{M \in \Mc} \KL{\proj_\Dc M^k}{M}
\)
with $M^k = M(\sv^k)$, we have
\begin{align}
    \sv^{k+1} = \argmin_\sv \, \Lc^\dagger(\sv, M^k), \label{eq:update_vk_SFBD}
\end{align}
where $\Lc^\dagger(\sv, M^k) = \KL{\proj_\Dc M^k}{M(\sv)}$. As detailed in \cref{appx:min_KL_eqv_drift_matching}, this is equivalent to minimizing
\begin{align*}
    \Lc(\sv, M^k) &= \textstyle \int_0^\tau \Lc_t \, \diff t 
    \numberthis \label{def:drift_matching_loss}
\end{align*}
where $\Lc_t := \Eb_{D(m_0^k)} \, \tfrac{1}{2} \left\| \tfrac{\xv_0 - \xv_t}{t} - \sv_t(\xv_t) \right\|^2$ and $m_0^k$ is the marginal of $M^k$ at $t = 0$. Thus, SFBD can be interpreted as the iterative update:
\begin{align}
    \sv^{k+1} = \argmin_\sv \, \Lc(\sv, M(\sv^k)),
    \label{eq:sfbd_s_update}
\end{align}
with $\sv^0 = \argmin_\sv \, \KL{\Dc(p_{\Ec_\text{clean}})}{M(\sv)}$. 

In practice, estimating each $\sv^k$ requires training a separate neural network, which is computationally expensive and hindered by manual intervention and ambiguous stopping criteria. While one could continue training from the previous step, the large shift in the target distribution $m_0^k$ across iterations forces optimizers such as Adam \citep{KingmaBa2014} to be reset, as stale momentum otherwise leads to divergence. This repeated reinitialization discards accumulated momentum and slows training. In \cref{sec:SFBD_flow}, we reformulate the update as a continuous steepest descent of $\Lc$, ensuring that $P^k$ evolves smoothly and $u_{\thetav}$ can be optimized without resetting, allowing the optimizer to adapt seamlessly.


\section{SFBD FLOW}
\label{sec:SFBD_flow}
In this section, we extend SFBD to a family of iterative deconvolution procedures, $\gamma$-SFBD with $\gamma \in (0,1]$. It recovers SFBD at $\gamma=1$, while as $\gamma \to 0$ the sequence ${M^k, \sv^k}$ converges to continuous flows ${M^\kappa, \sv^\kappa}$, $\kappa \geq 0$. 

We show that $\gamma$-SFBD admits two derivations: a generalized D-Proj, showing how smaller $\gamma$ yields smoother trajectories, and a discretized functional gradient descent on $\Lc(\sv, M_0(\sv))$, establishing convergence to a continuous flow. 

\textbf{Derive $\gamma$-SFBD Through a Generalized D-Proj.} For $\gamma \in (0,1]$, consider a generalized D-Proj:
\begin{align*}
 P^{k+1, \gamma}\!\! = \!\underset{P \in \Dc}{\argmin} \, (1\! -\! \gamma) \KL{P^{k, \gamma}}{P}\! + \!\gamma\KL{M^k}{P} \numberthis \label{def:gamma_diff_proj}
\end{align*}

We refer to SFBD with D-Proj replaced by $\gamma$-D-Proj as $\gamma$-SFBD. When $\gamma = 1$, it recovers the original SFBD. (Although $M^k$ does depend on $\gamma$, we keep the original notation for simplicity.) To see how $\gamma$-D-Proj smooths the update, note that the denoised samples at iteration $k$ follow a distribution with density (see \cref{appx:opt_sol_gamma_diff_proj}):
\begin{align}
    p_0^{k+1, \gamma} = (1 - \gamma)\, p_0^{k, \gamma} + \gamma\, m_0^k \label{eq:gamma_diff_proj_t=0}
\end{align}
where $p^{0,\gamma}_0 = p_{\Ec_\text{clean}}$ and $P^{k+1,\gamma} = D(p_0^{k+1, \gamma})$. Basically, the parameter $\gamma$ controls how much of the denoised set is updated using the latest model. When $\gamma = 1$, all samples are replaced, recovering standard SFBD. 
As $\gamma \to 0$, the updates become infinitesimal, leaving $M^{k+1}$ -- obtained by projecting $P^{k+1,\gamma}$ onto $\Mc$ -- and its corresponding $\sv^{k+1}$ nearly unchanged (see \cref{fig:sfgd_proj}b). 
Despite the smoothing effect, $\gamma$-SFBD guarantees convergence for all $\gamma \in (0,1]$. In particular, let $\Phi_p(\uv) = \Eb_p[\exp(i\, \uv^\top \xv)]$ denote the characteristic function of $p$ for $\uv \in \Rb^d$. Under mild assumptions, 
\begin{restatable}{proposition}{convGammaSFBD}
\label{prop:convGammaSFBD}
    For $k \geq 0$,
    \(
        \KL{p_{\rm data}}{p_0^{k\hspace{-0.05em}+\hspace{-0.05em}1, \gamma}} \hspace{-0.12em}  - \hspace{-0.12em} \KL{p_{\rm data}}{p_0^{k, \gamma}} \hspace{-0.12em} \leq \hspace{-0.12em} -\gamma  \KL{p_\tau^*}{p_\tau^{k, \gamma}}
    \). 
    In~addition, 
   	\begin{equation*}
	\min_{k = 1,\ldots, K} \left| \Phi_{p_{\rm data}}(\uv) - \Phi_{p_0^{k, \gamma}}(\uv) \right| 
	\leq \exp\Big(\tfrac{\tau}{2}\|\uv\|^2\Big) 
	\Big(\tfrac{2M}{\gamma K}\Big)^{\sfrac12}
	\end{equation*}
	for $K \geq 1,\, \uv \in \Rb^d$, and $M = \KL{p_{\rm data}}{p_{\Ec_{\rm clean}}}$. 
\end{restatable}
(All proofs are deferred to the appendix.) \cref{prop:convGammaSFBD} shows that for any $\gamma \in (0,1]$, $p_0^{k,\gamma}$ converges to $p_{\rm data}$ as $k$ grows, with convergence of characteristic functions ensuring convergence of distributions. Notably, although we report convergence speed with a fixed $\gamma$ for clarity, $\gamma \in (0,1]$ may vary across iterations without compromising convergence to the optimal solution. In \cref{prop:convGammaSFBDInperfect}, we show how to use an adaptive $\gamma$ to accommodate the imperfect estimation of neural networks.

\textbf{$\gamma$-SFBD as Functional Gradient Descent.} In \cref{sec:SFBD_alt_proj}, we showed that SFBD updates the backward drift $\sv^k$ by solving $\argmin_{\sv}~\Lc(\sv, M_0(\sv^k))$. We now consider a relaxed version, where $\sv$ is updated via a single gradient descent step in function space with step size $\gamma \in (0,1]$. This update rule exactly recovers $\gamma$-SFBD algorithm.

Recall that for a functional $\ell: \Fc \to \Rb$ defined over a function space $\Fc$, its functional derivative at $\uv \in \Fc$ with respect to a reference measure $P$ is a function $\nabla_P \ell(\uv) \in \Fc$ (when it exists) satisfying~\citep{CourantH1989}:
\begin{align*}
\textstyle \int\hspace{-0.3em} \inner{\nabla_P \ell(\uv)(\xv)}{\nuv(\xv)}\hspace{-0.3em} \diff \mu(\xv)\hspace{-0.2em}=\hspace{-0.2em} \lim_{\lambda \rightarrow 0}\hspace{-0.15em} \frac{1}{\lambda} \big( \ell(\uv + \lambda \nuv)\hspace{-0.15em} - \ell(\uv) \big)
\end{align*}
for all $\nuv \in \Fc$. Building on this, we have:
\begin{restatable}{proposition}{NonParaUpdateV}
\label{prop:non_para_update_v}
Let $\gamma \in (0, 1]$ and $k \in \Nb$. Let $P^{k, \gamma}$ and $M^k$ denote the stochastic process sequences generated by $\gamma$-SFBD via the update rules in \cref{def:markov_proj} and \cref{def:gamma_diff_proj}. Then the update of $M(\sv^k) = M^k$ satisfies
\begin{align}
\sv_t^{k+1}(\xv) = \sv_t^k(\xv) - \gamma \nabla_{P_t^{k+1, \gamma}} \Lc_t(\sv_t^k, m_0(\sv^k))(\xv) \label{eq:drift_func_grad_descent}
\end{align}
for all $\xv \in \Rb^d$ and $t \in [0, \tau]$.
\end{restatable}
As a result, $\gamma$-SFBD basically performs a discretized functional gradient descent on $\Lc(\sv, M_0(\sv))$ with step size $\gamma$, following the steepest descent under the reference distribution $P^{k,\gamma}$, updated via~\eqref{eq:gamma_diff_proj_t=0}. Remarkably, \cref{prop:convGammaSFBD} shows that value $\gamma$ does not affect convergence of $p_0^{k, \gamma}$ to $p_{\rm data}$. Thus, \textit{for any $\gamma \in (0,1]$, $\sv_t^k$ converges to the true score function $\sv_t^*$ learned by a diffusion model trained on clean data, with $\gamma = 1$ recovering the original SFBD result \citep{LuWY2025}}.

\textbf{SFBD Flow.} The functional gradient descent perspective shows that as $\gamma \to 0$, the discrete sequence $\{\sv^k\}_{k \in \Nb}$ and the associated distributions $p_0^{k,\gamma}$ converge to continuous flows $\{\sv^\kappa\}_{\kappa \geq 0}$ and $p_0^\kappa$, governed by the gradient flow of $\Lc(\sv, M_0(\sv))$. We refer to this continuous formulation as \emph{SFBD flow}. To characterize the evolution of $p_0^\kappa$, fix $\kappa > 0$ and let $\{\gamma_i\} \to 0$ with $k_i = \kappa/\gamma_i \in \Nb$. Then $p_0^{k_i, \gamma_i} \to p_0^\kappa$ and $m_0^{k_i} \to m_0(\sv^\kappa)$ via Euler approximation. Taking the limit,
\begin{align*}
\frac{\diff}{\diff \kappa} p_0^\kappa
=& \lim_{i \rightarrow \infty} \tfrac{1}{\gamma_i}(p_0^{k_i+1, \gamma_i} - p_0^{k_i, \gamma_i}) \overset{\eqref{eq:gamma_diff_proj_t=0}}{=}  m_0(\sv^\kappa) - p_0^\kappa,
\end{align*}
where $p^0_0 = p_{\Ec_\text{clean}}$. Thus, $p_0^\kappa$ evolves according to an ODE driven by the mismatch between the model's denoised output $m_0(\sv^\kappa)$ and the current estimate $p_0^\kappa$. Under this flow formulation, the convergence of $\gamma$-SFBD reduces to:
\begin{restatable}{corollary}{convCtnSFBD}
\label{cor:cont_flow_convergence}
For $\kappa > 0$, we have
\(
\frac{\diff}{\diff \kappa} \KL{p_{\rm data}}{p_0^\kappa} \leq -\KL{p_\tau^*}{p_\tau^\kappa}
\). Additionally, 
\begin{align*}
\textstyle \inf_{\kappa \in [0, \Kc]} \left|\Phi_{p_\text{data}}(\uv) - \Phi_{p_0^{\kappa}}(\uv)\right| \leq \exp\Big(\tfrac{\tau}{2}\|\uv\|^2\Big) \;\big(\frac{2 M}{\Kc}\big)^{\sfrac12}
\end{align*}
for  $\Kc > 0$, $\uv \in \Rb^d$ and $M = \KL{p_{\rm data}}{p_{\Ec_{\rm clean}}}$.
\end{restatable}

\textbf{Adaptive Step Size $\gamma$.} In practice, $\sv^k_t$ is approximated by a neural network $\hat\sv^{k}_t$. The imperfect approximation may hinder convergence to the true data distribution as empirically shown in \cref{sec:emp}. In \cref{prop:convGammaSFBDInperfect}, we show the issue can be mitigated by adaptively updating $\gamma$ using an upper bound $\delta_k$ on the approximation error, defined by
\begin{align}
\delta_k > \max\!\Big\{0,~\KL{P^\ast}{\hat M_k} - \KL{P^\ast}{M_k}\Big\}, \label{eq:def_deltak}
\end{align}
where $\hat M_k \in \Mc$ denotes the path measure induced by the approximate drift $\hat\sv^{k}_t$.
\begin{restatable}{proposition}{convGammaSFBDInperfect}
\label{prop:convGammaSFBDInperfect}
Suppose $\sv_t^k$ is approximated by a neural network $\hat\sv_t^{k}$, and let $\delta_k$ satisfy \cref{eq:def_deltak} with $\sum_{k=1}^\infty \delta_k < \infty$.  
For $\rho, \Gamma \in (0,1)$ and $\gamma_0, v_0 > 0$, choose the step size at iteration $k$ as
\begin{align}
	\gamma_k = \min\!\left(\tfrac{\gamma_0}{\sqrt{v_k}}, \Gamma \right), 
	\quad v_k = \rho v_{k-1} + (1-\rho)\delta_k^2.
\end{align}
Then $p^k_0 \to p_\text{data}$ as $k \to \infty$. 
\end{restatable}
\cref{prop:convGammaSFBDInperfect} basically suggests using larger step sizes $\gamma_k$ when the exponential moving average (EMA) of the error bound decreases. Intuitively, this means if the network $\hat\sv^k$ provides a good approximation of the target, a larger fraction of denoised samples can be substituted with those generated by $\hat\sv^k$ without hindering convergence. We provide a practical way to construct $\delta_k$ in \cref{sec:online_SFBD}. 

The parameter $\Gamma$ acts as an upper bound on the fraction of samples in $\mathcal{E}$ that may be refreshed at each iteration. Early in training, $\delta_k$ is typically large, which yields a large $v_k$ and therefore a very small effective update ratio $\frac{\gamma_0}{\sqrt{v_k}}$. In this stage, $\Gamma$ is essentially inactive. As training progresses and the model better aligns with the distribution represented by $\mathcal{E}$, $\frac{\gamma_0}{\sqrt{v_k}}$ may grow large. At this point, $\Gamma$ becomes the active constraint, limiting the maximum update ratio and preventing overly aggressive updates.

\section{ONLINE SFBD OPTIMIZATION}
\label{sec:online_SFBD}



As discussed in \cref{sec:SFBD_flow}, when $\gamma$ is small, \cref{prop:non_para_update_v} shows that the sequence ${\sv^k}$ closely tracks its continuous limit $\sv^\kappa$. Since $\sv^k$ is parameterized by neural networks, this continuity motivates replacing iterative fine-tuning in SFBD with a single network $\sv^{\phiv}$ that continuously approximates the evolving $\sv^k$. The optimization of $\sv^{\phiv}$ follows M-Proj~\cref{def:markov_proj}, implemented by minimizing the loss of matching score~\cref{eq:denoiser_loss} with $p_0 = p_0^{k, \gamma}$. Unlike standard SFBD, $\gamma$-SFBD refreshes only a fraction $\gamma$ of denoised samples in each $\gamma$-D-Proj step, inducing small changes to $p_0^{k, \gamma}$ - so a few gradient steps suffice for $\sv^{\phiv}$ to track the new minimizer. Building on this insight, we propose \textit{Online SFBD} in \cref{alg:Online_SFBD}, which eliminates the need to fine-tune a sequence of networks.

\textbf{Adaptive step size $\gamma$.} In \cref{prop:convGammaSFBDInperfect}, we show that the imperfect approximation of network $\hat\sv_t$ can be accommodated by adjusting the step size, which requires us to build valid $\delta_k$ satisfying \cref{eq:def_deltak}. In our implementation, we construct $\Ec_{\phiv}$ by sampling 256 images $\yv$ from $\Ec_\text{noisy}$ and solving~\cref{eq:anderson_bwd} from $t=\tau$ to $0$ with drift $\hat\sv_t(\xv_t)$ and initial condition $\xv_\tau=\yv$. We then heuristically set $\delta^2 \propto \max\big(0,\text{\sffamily KID}(\Ec_\text{clean}, \Ec_{\phiv}) - \text{\sffamily KID}(\Ec_\text{clean}, \Ec)\big)$, with the scaling absorbed into $\gamma_0$. Here $\text{\sffamily KID}$ denotes Kernel Inception Distance~\citep{BinkowskiSAG2018}, an unbiased alternative to Fréchet Inception Distance~\citep{HeuselRUNH2017} that does not require a minimum sample size. Given $\delta^2$, \cref{alg:gamma_function} defines the update of $\gamma$. If $\delta^2 \geq \eta v$, we set $\gamma=0$ and skip the partial update of $\Ec$, allowing $\hat\sv$ more steps to converge before the target shifts. This design ensures that the effective $\delta$ decays faster than a geometric sequence with some ratio $\alpha < 1$, guaranteeing $\sum_k \delta_k < \infty$ as required by \cref{prop:convGammaSFBDInperfect} (see \cref{appx:delta_k_geometric_decay} for details and justification of the chosen $\delta^2$ formulation).

{
\setlength{\belowcaptionskip}{0.1em} 
\setlength{\textfloatsep}{0.1em}

\begin{algorithm}[t]
\LinesNotNumbered
\DontPrintSemicolon
   \caption{Online SFBD}
   \label{alg:Online_SFBD}

    \textbf{Input:} clean data: $\Ec_\text{clean} = \{\xv^{(i)}\}_{i=1}^M$, noisy data: $\Ec_\text{noisy} = \{\yv_\tau^{(i)}\}_{i=1}^N$, num of gradient steps: $m$, base $\gamma_0$
    
    \tcp{Initialize Denoiser}
    
    $\phiv \leftarrow$ Pretrain $D_{\phiv}$ using \cref{eq:denoiser_loss} with $p_0 = p_{\Ec_\text{clean}}$     \vspace{-0.3em}
   
    $\Ec \leftarrow \{ \yv_0^{(i)}:$ solve \cref{eq:anderson_bwd} from $t = \tau$ to $0$ with  $\sv_t(\xv_t) = \hat \sv_t(\xv_t) := \tfrac{D_{\phiv}(\xv_t, t) - \xv_t}{t}, \; \xv_\tau = \yv_\tau^{(i)} \in \Ec_\text{noisy} \}$
    
	$v \leftarrow$ compute $\delta$ for $\hat \sv_t(\xv_t)$ \textbf{if} adapt. $\gamma$, \textbf{else} $-1$
    
    \Repeat{reach the maximum number of iterations}{       
        Update $\phiv$ with $m$ gradient steps on \cref{eq:denoiser_loss} with $p_0 = p_\Ec$.\hfill \tcp{M-Proj}
        
        $\delta \leftarrow$ compute $\delta$ for $\hat \sv_t(\xv_t)$ \textbf{if} adapt. $\gamma$, \textbf{else} $-1$
        
        $(\gamma, v) \leftarrow$ \textsf{getStepSize}($v, \delta$) \textbf{if} adapt. $\gamma$, \textbf{else} $(\gamma_0, -1)$.
        
        $\Ec \leftarrow$ \{Replace ratio $\gamma$ of denoised samples in $\Ec$ with the new ones by solving~\cref{eq:anderson_bwd} from $t = \tau$ to $0$ with  $\sv_t(\xv_t) = \hat\sv_t(\xv_t) := \tfrac{D_{\phiv}(\xv_t, t) - \xv_t}{t}, \;\xv_\tau = \yv_\tau^{(i)}$ randomly picked from $\Ec_\text{noisy}$\} \hfill \tcp{$\gamma$-D-Proj}
    }
    \KwOut{Final denoiser $D_{{\phiv}}$}
\end{algorithm}
}
{\begin{algorithm}[t]
\DontPrintSemicolon
\LinesNotNumbered
   \caption{\textsf{getStepSize} -- Adaptive step size update}
   \label{alg:gamma_function}

   \KwIn{EMA $v$, error bound $\delta$, threshold $\eta \in (0,1)$, decay $\rho \in (0,1)$, base $\gamma_0 > 0$, cap $\Gamma \in (0,1]$.}
   
   \eIf{$\delta^2 < \eta \, v$}{
       $v \leftarrow \rho \, v + (1-\rho)\,\delta^2$ \tcp*{Update EMA}
       
       $\gamma \leftarrow \min\!\big(\tfrac{\gamma_0}{\sqrt{v}},\, \Gamma\big)$ \tcp*{Compute step size}
   }{
       $\gamma \leftarrow 0$ \tcp*{Skip $\Ec$ update for large error}
   }
   \KwOut{$(\gamma, v)$}
\end{algorithm}}

\textbf{Combining Denoised and Clean Samples.} Since the copyright-free clean samples are drawn from the true data distribution, we follow the original SFBD framework \citep{LuWY2025} and set $p_0 = p_{\Ec \cup \Ec_\text{clean}}$ in the M-Proj step. This choice helps accelerate optimization by aligning the target distribution for updating $\phiv$ more closely with the true data distribution. As detailed in \cref{appx:gamma_SFBD_v2}, this corresponds to a variant of $\gamma$-Diff Proj, and we provide additional justification there for the observed performance gains.

\textbf{Denoising and Sampling.} 
While \cref{alg:Online_SFBD} uses a naive backward sampler by solving \cref{eq:anderson_bwd}, the algorithm allows any backward SDE and solver that yield the same marginals. We adopt the 2\textsuperscript{nd}-order Heun method  from EDM~\citep{KarrasAAL22} for better error control and efficiency. To improve sample quality~\citep{NicholD2021, KarrasAAL22}, we maintain an EMA version of the model for denoising and use it to update $\Ec$; all reported results in \cref{sec:emp} are based on this EMA model. In practice, $\gamma$ is typically small (e.g., $\gamma < 0.02$), so the mild asynchrony between $\gamma$-D-Proj and M-Proj has negligible effect, as suggested by preliminary exploration during framework implementation. This motivates a practical strategy we call \textit{asynchronous denoising}: denoising runs independently on a separate, low-performance GPU, updating $\Ec$ in the background, while the main training loop minimizes \cref{eq:denoiser_loss} on high-performance hardware using the latest $p_0 = p_\Ec$. We adopt this strategy throughout our study in \cref{sec:emp}.

\textbf{Relationship to Consistency Constraint-based Methods}. 
Consistency constraint-based (CC-based) methods such as TweedieDiff~\citep{DarasDD2024} and TweedieDiff+~\citep{DarasCD25}, which enforce consistency only between time zero and positive time steps, can be seen as special cases of Online SFBD with a single gradient step ($m = 1$). (See \cref{appx:rel_constency_based_mtd} for details and an extension to arbitrary time pairs.) These methods approximate $p^{k, \gamma}$ using $m_0^k$ rather than the EMA over $\{m_0^j\}_{j \leq k}$ as defined in \cref{eq:gamma_diff_proj_t=0}, which is not exact unless $\gamma = 1$. Since $\sv$ is updated just once per iteration, $m_0^j$ for $j$ close to $k$ tends to be similar, making $m_0^k$ a reasonable proxy of $p^{k, \gamma}$ when $\gamma$ is not too small.

In \cref{sec:emp}, we show that avoiding this approximation enables Online SFBD to consistently outperform CC-based methods. Remarkably, it also achieves significantly lower computational cost. This is because Online SFBD reuses cached denoised samples throughout training, whereas CC-based methods generate them on demand -- requiring more samples per step for stability and making asynchronous denoising impractical. Moreover, CC-based methods typically enforce consistency between arbitrary time pairs, requiring multiple neural network forward passes per update. In contrast, Online SFBD matches the compute cost of a standard diffusion model, apart from denoised sample updates -- which can be performed asynchronously on separate GPUs.

\section{EMPIRICAL STUDY}
\label{sec:emp}

\begin{figure*}
\centering     
\subfloat[Clean Image Ratio (fixed $\gamma$)]{\label{fig:ablation_cifar10:clean_ratio}\includegraphics[height=4.1cm]{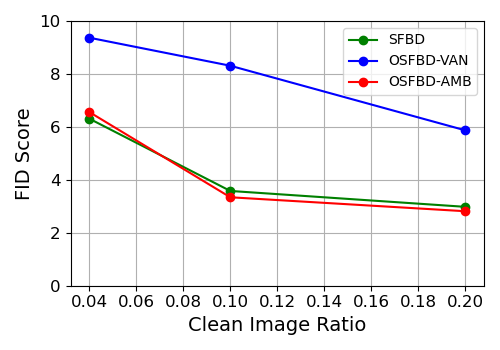}}\hfill
\subfloat[Noise Level (fixed $\gamma$)]{\label{fig:ablation_cifar10:sigma}\includegraphics[height=4cm]{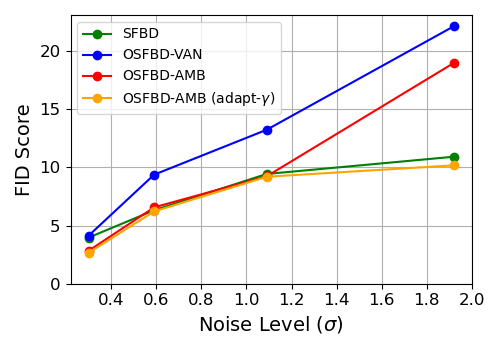}}\hfill
\subfloat[Num of Grad. Steps per Iteration]{\label{fig:ablation_cifar10:pretrain}\includegraphics[height=3.85cm]{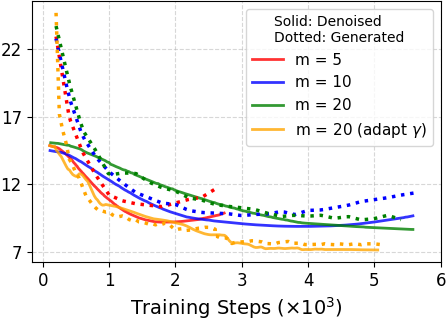}}
\caption{FID scores of Online SFBD (OSFBD) on CIFAR-10 under different settings. Unless specified, the clean ratio is $0.04$, noise level $\sigma=0.59$, and gradient steps $m=20$.  (c) reports results with OSFBD-VAN pretraining. 
}
\label{fig:ablation_cifar10}
\end{figure*}


In this section, we demonstrate the effectiveness of Online SFBD. We first study its behaviour under different configurations to identify practical settings, with ablation results supporting the theory and offering guidance for use. Building on these insights, we benchmark Online SFBD and show that it consistently outperforms models trained directly on noisy data. Compared to standard SFBD, the online variant achieves better results while eliminating costly iterative finetuning and denoising.

\textbf{Datasets and Evaluation Metrics.} We conduct experiments on CIFAR-10~\citep{Krizhevsky2009} and CelebA~\citep{LiuGL2022}, using image resolutions of $32 \times 32$ and $64 \times 64$, respectively. CIFAR-10 contains 50,000 training and 10,000 test images across 10 classes. CelebA includes 162,770 training, 19,867 validation, and 19,962 test images; we use the preprocessed version from the official DDIM repository~\citep{SongME2021}.  Corrupted images are generated by adding independent Gaussian noise with standard deviation $\sigma$ to each pixel after rescaling to $[-1, 1]$. Notably, only one noisy counterpart is generated per clean image.

Image quality is evaluated using Fréchet Inception Distance (FID), computed between the reference dataset and 50,000 model-generated samples. Generated image samples are shown in \cref{appx:sample_results}.

\textbf{Models and Other Configurations.} We implement Online SFBD using the EDM architecture and hyperparameters~\citep{KarrasAAL22} in an unconditional setting, with non-leaky augmentation to mitigate overfitting. Backward sampling is performed with the 2\textsuperscript{nd}-order Heun method~\citep{KarrasAAL22}; details are in \cref{appx:expConfig}. As discussed in \cref{sec:online_SFBD}, Online SFBD minimizes the denoising score-matching loss~\cref{eq:denoiser_loss} with $p_0 = p_0^{k,\gamma}$ updated via sample denoising. Effective training requires sufficiently minimizing \cref{eq:denoiser_loss}$,$ before each $p_0$ update. For fixed $\gamma$, this balance can be controlled either by adjusting the update ratio $\gamma$ or the number of gradient steps $m$. Since batch denoising is more efficient with larger updates, we fix $\gamma$ (updating 640 samples per iteration) and vary $m$, using $m=20$ by default unless otherwise noted. For adaptive $\gamma$, we fix $m=20$ and compute $\gamma$ using \cref{alg:gamma_function} with $\eta=0.99$ and $\rho=0.9$. The number of updated samples per $\gamma$-\textsf{D-Proj} step is capped at 2048, corresponding to $\Gamma=0.04$ for CIFAR-10 and $0.01$ for CelebA.

\textbf{Role and Practical Setting of $\Gamma$}. As discussed in \cref{sec:SFBD_flow}, $\Gamma$ becomes active only in the later stages of training, where it caps the maximum update ratio and prevents overly aggressive updates. Empirically, this cap slows optimization but improves stability.

As we will show in \cref{fig:ablation_cifar10}c, with a fixed $\gamma$, a small $m$ leads to a large effective update rate, yielding a rapid initial FID drop, whereas a smaller update rate slows progress yet stabilizes convergence. Considering that the adaptive-$\gamma$ scheme automatically adjusts the effective update ratio to prevent training collapse, in practice, we simply set $\Gamma$ to the value corresponding to the largest single sampling batch size that fits within GPU memory, which yields the fastest convergence. Following this principle, low-resolution datasets such as CIFAR-10 allow for a larger $\Gamma$ and therefore potentially faster convergence, whereas high-resolution datasets like CelebA require smaller $\Gamma$ values due to reduced batch capacity.

\subsection{Ablation Study}
\label{sec:ablation}
We examine the behaviour of Online SFBD under various settings on CIFAR-10. Informed configurations lead to state-of-the-art results on all benchmarks, as shown in \cref{sec:benchmark}.

\textbf{Methods of Pretraining.} While the score function for $t < \tau$ must be estimated from limited clean, copyright-free data, ambient score matching (ASM) can guide score estimation for $t > \tau$ using noisy samples~\citep{DarasCD25}. In \cref{fig:ablation_cifar10}, we compare models pretrained only on clean data (\textbf{OSFBD-VAN}\,ILLA) with those jointly pretrained using ASM and noisy samples (\textbf{OSFBD-AMB}\,IENT). As shown in (a) and (b), OSFBD-AMB consistently outperforms OSFBD-VAN across all clean image ratios and noise levels. This improvement is expected, as OSFBD-AMB better leverages the noisy data to refine score estimation for $t > \tau$ and boost overall performance. We therefore use OSFBD-AMB for the benchmarks in \cref{sec:benchmark}.

\begin{table}[t]
    \centering
    \begin{minipage}[t]{\linewidth}
        \centering
        \caption{Comparison. For $\sigma > 0$, models are trained on images corrupted with Gaussian noise $\mathcal{N}(\zero, \sigma^2 \Iv)$.}
        \label{tb:performance_compare}
        \resizebox{\columnwidth}{!}{%
            \begin{tabular}{@{}lccc|ccc@{}}
                \toprule
                \multirow{2}{*}{Method} 
                & \multicolumn{3}{c|}{CIFAR-10 (32×32)} 
                & \multicolumn{3}{c@{}}{CelebA (64×64)} 
                \\ \cmidrule(lr){2-4} \cmidrule(l){5-7}
                & $\sigma$ & \makecell{Pretrain\\(50 imgs)} & FID 
                & $\sigma$ & \makecell{Pretrain\\(50 imgs)} & FID 
                \\ \midrule
                DDPM~\citep{HoJA2020}              & 0.0 & No  & 4.04   & 0.0 & No  & \textbf{3.26} \\
                DDIM~\citep{SongME2021}           & 0.0 & No  & 4.16   & 0.0 & No  & 6.53          \\
                EDM~\citep{KarrasAAL22}           & 0.0 & No  & \textbf{1.97} & --  & --  & --           \\ \midrule
                EMDiff~\citep{BaiWCS2024}         & 0.2 & Yes & 86.47  & --  & --  & --           \\
                TweedieDiff~\citep{DarasDD2024}   & 0.2 & Yes & 65.21 & 0.2 & Yes & 58.52 \\
                TweedieDiff+~\citep{DarasCD25}    & 0.2 & Yes & 8.05 & 0.2 & Yes & 6.81 \\
                SFBD~\citep{LuWY2025}             & 0.2 & Yes & 13.53 & 0.2 & Yes & 6.49 \\
                OSFBD-AMB (fixed $\gamma$)                  & 0.2 & Yes & 3.22   & 0.2 & Yes & 3.23 \\
                	OSFBD-AMB (adaptive $\gamma$)                  & 0.2 & Yes & \textbf{3.12}   & 0.2 & Yes & \textbf{3.19} \\
                \bottomrule
            \end{tabular}
        }
    \end{minipage}
    \vspace{0.1em}
    
	\begin{minipage}[t]{\linewidth}
        \centering
        \caption{Additional results for competitive models under various settings. (All models are pretrained.)}
        \label{tb:performance_compare_addition}
        \resizebox{\columnwidth}{!}{%
            \begin{tabular}{@{}lccc|ccc@{}}
                \toprule
                \multirow{2}{*}{Method} 
                & \multicolumn{3}{c|}{CIFAR-10 (32×32)} 
                & \multicolumn{3}{c@{}}{CelebA (64×64)} 
                \\ \cmidrule(lr){2-4} \cmidrule(l){5-7}
                & $\sigma$ & clean samples & FID 
                & $\sigma$ & clean samples & FID 
                \\ \midrule
                TweedieDiff+~\citep{DarasCD25}   & 0.2 & 10\% & 2.81 & 1.38 & 50 & 35.65 \\
                SFBD~\citep{LuWY2025}           & 0.2 & 10\% & 2.58 & 1.38 & 50 & 23.63 \\
                OSFBD-AMB (fixed $\gamma$)                & 0.2 & 10\% & 2.73   & 1.38 & 50 & 27.09\\
                OSFBD-AMB (adaptive $\gamma$)                & 0.2 & 10\% & \textbf{2.49}   & 1.38 & 50 & \textbf{20.21}\\ \midrule
                TweedieDiff+~\citep{DarasCD25}   & 0.59 & 4\% & 6.75 & 1.38 & 1,500 & 6.81 \\
                SFBD~\citep{LuWY2025}           & 0.59 & 4\% & 6.31 & 1.38 & 1,500 & 5.91 \\
                OSFBD-AMB (fixed $\gamma$)                & 0.59 & 4\% & 6.56   & 1.38 & 1,500 & 5.72 \\
                OSFBD-AMB (adaptive $\gamma$)                & 0.59 & 4\% & \textbf{6.21}   & 1.38 & 1,500 & \textbf{5.40} \\
                \bottomrule
            \end{tabular}
        }
    \end{minipage}
    \vspace{-1em}
\end{table}
\textbf{Number of Gradient Steps and Adaptive $\gamma$.} \Cref{fig:ablation_cifar10:pretrain} shows the FID trajectories of generated and denoised samples during training for different gradient step counts $m$ for fixed and adaptive $\gamma$. The models are pretrained using OSFBD-VAN. 

The FID of denoised samples reflects the distance between $p^{k, \gamma}$ and $p_\text{data}$. Since a model that fully learns $p^{k, \gamma}$ would generate samples with FIDs matching the denoised ones, the FID gap indicates how well \cref{eq:denoiser_loss} has been minimized. A large gap suggests incomplete minimization at the current step. As shown in \Cref{fig:ablation_cifar10:pretrain}, smaller $m$ values lead to more frequent updates of $p^{k, \gamma}$, causing a faster initial FID drop for denoised samples (e.g., $m = 5$). However, this rapid updating prevents the model from fully learning $p^{k, \gamma}$ before it shifts, as indicated by the widening FID gap after 1.5k steps. Consequently, the denoising process degrades, and FIDs for both denoised and generated samples begin to rise. With larger $m$, the model has more time to minimize the loss before $p^{k, \gamma}$ changes, delaying such degradation and achieving lower FIDs overall. Similar trends are observed with OSFBD-AMB, though to a milder degree. Importantly, although a bigger $m$ can improve the training performance, this does not imply that an arbitrarily large $m$ should be chosen, as it will significantly slow down the training process. In practice, we find $m = 20$ strikes a good balance.


Adaptive $\gamma$ adjusts the update speed according to training progress, enabling faster FID reduction without the divergence seen at small $m$. The orange curve in \cref{fig:ablation_cifar10}c appears less smooth than with fixed $\gamma$ due to these adaptive adjustments, but this reflects the method’s responsiveness and results in stable, accelerated convergence. The close overlap of the orange solid and dotted lines further indicates that adaptive $\gamma$ provides enough iterations for the model to converge to each evolving target distribution.

\subsection{Performance Comparison}
\label{sec:benchmark}

We compare OSFBD-AMB with representative models for training on noisy images, as summarized in \cref{tb:performance_compare}. EMDiff~\citep{BaiWCS2024} uses a diffusion-based EM algorithm for inverse problems. TweedieDiff~\citep{DarasDD2024} applies the original consistency loss from \cref{eq:consistency_loss} and is pretrained on clean data. TweedieDiff+~\citep{DarasCD25} adopts the same pretraining as OSFBD-AMB, followed by joint training with a simplified consistency objective. SFBD~\citep{LuWY2025} is the original algorithm requring iteratively finetuning. 

\textbf{Benchmark}. Following the setup of \citet{BaiWCS2024, LuWY2025}, we use 50 clean samples along with data corrupted Gaussian noise ($\sigma = 0.2$), with the same clean set across all experiments. For reference, we also report results for models trained on fully clean data ($\sigma = 0$). As shown in \cref{tb:performance_compare}, OSFBD-AMB with fixed and adaptive $\gamma$ consistently outperforms all baselines, producing significantly higher-quality images. Notably, it even surpasses DDPM and DDIM trained exclusively on clean samples on both datasets.

To further evaluate OSFBD-AMB, we test it under additional dataset configurations (\cref{tb:performance_compare_addition}) against the two strongest baselines, TweedieDiff+ and SFBD. With fixed $\gamma$, OSFBD-AMB consistently outperforms TweedieDiff+ and matches SFBD in most cases, except on a challenging CelebA setting with scarce clean data and high noise ($\sigma$). The adaptive $\gamma$ variant delivers consistently superior performance. 

\textbf{OSFBD-AMB vs SFBD.} Results in \cref{tb:performance_compare,tb:performance_compare_addition} show that SFBD can outperform fixed-$\gamma$ OSFBD-AMB in cases with very limited clean data and high noise. To probe this further, we compare SFBD and OSFBD-AMB on CIFAR-10 across the same clean-data ratios and noise levels used in the original SFBD setup~\citep{LuWY2025} (see \cref{fig:ablation_cifar10}a,b). Specifically, when a moderate amount of clean data is available, the two methods perform comparably. Under low noise and scarce clean data, OSFBD-AMB outperforms SFBD, likely due to ambient-based pretraining and smoother updates of $p^{k,\gamma}$, which mitigate the overfitting issues noted in SFBD on small datasets~\citep{LuWY2025}. At high noise levels, however, denoising requires more backward SDE steps, compounding errors from imperfect training. In this regime, accurate score estimation becomes critical and demands many more gradient steps $m$ for fixed-$\gamma$ OSFBD-AMB, making SFBD more stable and effective. By contrast, adaptive-$\gamma$ OSFBD-AMB tracks estimation quality and dynamically adjusts step size, alleviating imperfect training and yielding the superior performance seen in \cref{tb:performance_compare,tb:performance_compare_addition}.

To demonstrate the superior training efficiency of the proposed flow-based method, \cref{tb:time_compare} presents a comparison of training times for SFBD and OSFBD-AMB (with adaptive $\gamma$) on CIFAR-10 and CelebA. As shown, SFBD requires approximately 50\% more training time than OSFBD-AMB. Notably, this difference does not yet account for the additional human intervention and iterative tuning required across SFBD runs.

\begin{table}[t]
\centering
\caption{Training time comparison between SFBD and OSFBD-AMB (adaptive $\gamma$) on CIFAR-10 and CelebA.}
\label{tb:time_compare}
\resizebox{\columnwidth}{!}{%
\begin{tabular}{@{}lccc|ccc@{}}
\toprule
\multirow{2}{*}{Method}
& \multicolumn{3}{c|}{CIFAR-10 ($\sigma=0.59$, 4\% clean)}
& \multicolumn{3}{c@{}}{CelebA ($\sigma=1.38$, 1500 clean)} \\
\cmidrule(lr){2-4} \cmidrule(l){5-7}
& Pretrain & Optimization & Total
& Pretrain & Optimization & Total \\
\midrule
SFBD & 24 h & $18$ h $\times 4$ & 96 h & 36 h & $24$ h $\times 5$ & 156 h \\
OSFBD-AMB & 24 h & 36 h & 60 h & 36 h & 72 h & 108 h \\
\bottomrule
\end{tabular}
}
\vspace{-1em}
\end{table}

\textbf{OSFBD-AMB vs TweedieDiff+.} We observe that OSFBD-AMB with adpative and fixed $\gamma$ consistently outperforms TweedieDiff+ across all settings in \cref{tb:performance_compare,tb:performance_compare_addition}, consistent with our discussion on their relationship in \cref{sec:online_SFBD}. Both methods share the same pretraining procedure and differ only in how they learn the score function for $t < \tau$. By updating the denoised sample set in an EMA-like manner, OSFBD-AMB presents a significantly more accurate target distribution $p^{k,\gamma}$, leading to improved performance. Notably, this improvement also reduces computational cost -- though at the expense of additional memory to cache denoised samples.

\section{DISCUSSION}
\label{sec:disc}
\vspace{-0.5em}
This paper extends the original SFBD algorithm to a family of variants, $\gamma$-SFBD. When $\gamma = 1$, it recovers SFBD; as $\gamma \to 0$, it yields SFBD flow and its practical counterpart -- Online SFBD -- which eliminates the need for alternating between denoising and fine-tuning. We also highlight its close connection to CC-based methods, another class of leading diffusion-based deconvolution techniques. Empirical results corroborate our analysis, showing that Online SFBD consistently outperforms strong baselines across some benchmarks.

\section*{Acknowledgement}
We gratefully acknowledge funding support from NSERC, the Canada CIFAR AI Chairs program and the Ontario Early Researcher program. Resources used in preparing this research were provided, in part, by the Province of Ontario, the Government of Canada through CIFAR, and companies sponsoring the Vector Institute.

%
%

\bibliography{deconv}
\bibliographystyle{plainnat}

\section*{Checklist}



\begin{enumerate}

  \item For all models and algorithms presented, check if you include:
  \begin{enumerate}
    \item A clear description of the mathematical setting, assumptions, algorithm, and/or model. [Yes] All assumptions for the given theorem are clearly stated if needed, with more detailed derivations provided in the appendix. We provide clear descriptions of the algorithms in \cref{alg:Online_SFBD} and \cref{alg:gamma_function}.
    \item An analysis of the properties and complexity (time, space, sample size) of any algorithm. [No] An analysis of algorithmic properties and complexity (time, space, sample size) is not applicable here, as this work does not focus on complexity. Nonetheless, we provide a general discussion of the memory requirements of our proposed algorithms relative to existing methods.
    \item (Optional) Anonymized source code, with specification of all dependencies, including external libraries. [No] We will provide anonymized source code if requested but not otherwise.
  \end{enumerate}

  \item For any theoretical claim, check if you include:
  \begin{enumerate}
    \item Statements of the full set of assumptions of all theoretical results. [Yes] We provide the full set of assumptions for each theoretical result.
    \item Complete proofs of all theoretical results. [Yes] We provide a complete proof of theoretical results in the appendix.
    \item Clear explanations of any assumptions. [Yes] We clearly explain all assumptions.
  \end{enumerate}

  \item For all figures and tables that present empirical results, check if you include:
  \begin{enumerate}
    \item The code, data, and instructions needed to reproduce the main experimental results (either in the supplemental material or as a URL). [No] The code is not currently provided but will be released upon acceptance. While the supplementary materials provides sufficient details, additional usage instructions will be provided with code release.
    \item All the training details (e.g., data splits, hyperparameters, how they were chosen). [Yes] Refer to \cref{appx:model_arch}
    \item A clear definition of the specific measure or statistics and error bars (e.g., with respect to the random seed after running experiments multiple times). [No] Training diffusion models from scratch is computationally expensive, and our access to high-performance hardware is limited in our school. Consequently, each configuration is run once, which aligns with common practice in diffusion model research. Nonetheless, the results are consistent across datasets and configurations, supporting the reliability of our conclusions.
    \item A description of the computing infrastructure used. (e.g., type of GPUs, internal cluster, or cloud provider). [Yes] Refer to \cref{appx:hardware_config}
  \end{enumerate}

  \item If you are using existing assets (e.g., code, data, models) or curating/releasing new assets, check if you include:
  \begin{enumerate}
    \item Citations of the creator If your work uses existing assets. [Yes] We reference the licenses of existing datasets and cite the original papers corresponding to all code used.
    \item The license information of the assets, if applicable. [Yes]
    \item New assets either in the supplemental material or as a URL, if applicable. [Not Applicable] No new assets were released.
    \item Information about consent from data providers/curators. [Not Applicable] No consent from data providers was required.
    \item Discussion of sensible content if applicable, e.g., personally identifiable information or offensive content. [Not Applicable] No sensible content was used.
  \end{enumerate}

  \item If you used crowdsourcing or conducted research with human subjects, check if you include:
  \begin{enumerate}
    \item The full text of instructions given to participants and screenshots. [Not Applicable]
    \item Descriptions of potential participant risks, with links to Institutional Review Board (IRB) approvals if applicable. [Not Applicable]
    \item The estimated hourly wage paid to participants and the total amount spent on participant compensation. [Not Applicable]
  \end{enumerate}

  Justification: The project does not use crowdsourcing nor did it conduct research with human subjects.

\end{enumerate}

\newpage
\appendix
\onecolumn
\section{THEORETICAL RESULTS}


\subsection{The Equivalence of the Forward Process Among the Existing Diffusion Models}
\label{sec:eqv_fwd_proc}

With some simple extension, our work can handle a more general forward diffusion process of the form 
\begin{align}
	\diff \xv_t=f(t)\xv_t\,\diff t + g(t)\,\diff \wv_t. 	\label{eq:fwd_diff_general}
\end{align}
It can be shown that its transition kernel is a Gaussian distribution such that
\begin{align}
	p(\mathbf{x}_t|\xv_0) = \Nc(m_t \mathbf{x}_0, \sigma_t^2 I)
\end{align}
See, for example, \citep[Appendix B.1]{KarrasAAL22} or \citep[Chapter 6]{SarkkaS2019}. To the best of our knowledge, almost all popular diffusion-based methods adopt a forward process that can be written of this general form. Based on this fact, starting from $\xv_0$,  $\xv_t$ can be seen as a sample of form
\begin{align}
	\xv_t = m_t \xv_0 + \sigma_t \epsilonv
\end{align}
where $\epsilonv \sim \Nc(\mathbf{0}, I)$. 
As a result, we obtain
\begin{align}
	\xv_t^\prime := \frac{\xv_t}{m_t} = {\xv}_0 + \frac{\sigma_t}{m_t} \epsilonv
\end{align}
which shows that $\xv_t'$ is equivalent to a noisy observation of $\xv_0$ corrupted by Gaussian noise with noise level $\frac{\sigma_t}{m_t}$. This implies that the more general formulation considered here naturally reduces to the setting studied in the main paper. Consequently, our proposed method can also be implemented using the forward process in \cref{eq:fwd_diff_general}, while the backward denoising step \cref{eq:anderson_bwd} becomes
\begin{align}
	\diff \xv_t=f(t)\xv_t \, \diff t - g(t)^2 \, \nabla\log p_t(\xv_t) + g(t) \, \diff \wv_t. 
\end{align}

\subsection{Minimizing KL Divergence is Equivalent to Conditional Drift Matching}
\label{appx:min_KL_eqv_drift_matching}
In \cref{sec:SFBD_alt_proj}, we claimed that minimizing $\Lc^\dagger$ defined in \cref{eq:update_vk_SFBD} is equivalent to minimizing
\begin{align}
	\Lc(\sv, M^k) = \int_0^\tau \Lc_t \diff t = \int_0^\tau {\Eb}_{D(m^k_0)} \, \tfrac{1}{2}\left\| \frac{\xv_0 - \xv_t}{t} - \sv_t(\xv_t)\right\|^2 \diff t. 
\end{align}
To see this, note that according to \cref{eq:disintegration_thm_D_Proj}, $D$-Proj sets $P^{k+1} = \proj_\Dc \, M^k = D(m_0^k)$. As a result, 
\begin{align*}
	\Lc^\dagger(\sv, M^k) &= \KL{\proj_\Dc \, M^k}{M(\sv)} = \KL{D(m_0^k)}{M(\sv)}. 
\end{align*}
By \cref{appx:lem:kl_two_path}, the KL divergence 
\begin{align*}
	\KL{D(m_0^k)}{M(\sv)} = \underbrace{\KL{m_0^k \conv \Nc(\zero, \tau \Iv)}{p^*_\tau}}_{\text{const.}} + \underset{D(m_0^k)}{\Eb} \int_0^\tau \frac{1}{2 } \left\|\bv(\xv_t, t) - \sv_t(\xv_t) \right\|^2  \diff t,
\end{align*}
where $\bv^k(\xv_t, t)$ is the drift of the backward SDE starting from $\tau$ with the initial distribution $m_0^k  \conv \Nc(\zero, \tau \Iv)$. \citet{anderson1982} showed that $\bv^k(\xv_t, t) =  \nabla \log m_t^k(\xv_t)$, where $m_t^k(\xv_t)$ denotes the density of the marginal distribution of $M^k$. It can be shown that (e.g., see \citep[Lemma 1]{SongDCS2023}):
\begin{align}
	\nabla \log m_t^k(\xv_t) = \Eb_{m^k_{0|t}}[\nabla_{\xv_t} \log m_t^k(\xv_t \vert \xv_0) \vert \xv_t ] = \Eb_{m^k_{0|t}}\left[\frac{1}{t}(\xv_0 - \xv_t) \Big\vert \xv_t\right]. 
\end{align}
As a result,
\begin{align*}
	&\underset{D(m_0^k)}{\Eb} \int_0^\tau \frac{1}{2 } \left\|\bv(\xv_t, t) - \sv_t(\xv_t) \right\|^2 \diff t = \underset{D(m_0^k)}{\Eb} \int_0^\tau \frac{1}{2 } \left\|\Eb_{m^k_{0|t}}\left[\frac{1}{t}(\xv_0 - \xv_t) \Big\vert \xv_t\right] - \sv_t(\xv_t) \right\|^2 \diff t. 
\end{align*}
Therefore,
\begin{align*}
	\argmin_\sv \tilde\Lc(\sv, M^k) &=  \argmin_\sv ~~ \underset{D(m_0^k)}{\Eb} \int_0^\tau\frac{1}{2 } \left\|\bv(\xv_t, t) - \sv_t(\xv_t) \right\|^2  \diff t\\
	&= \argmin_\sv \underset{D(m_0^k)}{\Eb} \int_0^\tau \left\|\Eb_{m^k_{0|t}}\left[\frac{1}{t}(\xv_0 - \xv_t) \Big\vert \xv_t\right] - \sv_t(\xv_t) \right\|^2 \diff t. 
\end{align*}
In addition, for any $t \in [0,\tau]$,
\begin{align*}
	&\underset{D(m_0^k)}{\Eb} \left\|\tfrac{1}{t}(\xv_0 - \xv_t) - \sv_t(\xv_t) \right\|^2\\
	=& \underset{D(m_0^k)}{\Eb} \left\|\tfrac{1}{t}(\xv_0 - \xv_t) - \Eb_{m^k_{0|t}}\left[\tfrac{1}{t}(\xv_0 - \xv_t) \Big\vert \xv_t\right] + \Eb_{m^k_{0|t}}\left[\tfrac{1}{t}(\xv_0 - \xv_t) \Big\vert \xv_t\right] - \sv_t(\xv_t) \right\|^2 \\
	=& \underset{D(m_0^k)}{\Eb} \left\|\tfrac{1}{t}(\xv_0 - \xv_t) - \Eb_{m^k_{0|t}}\left[\tfrac{1}{t}(\xv_0 - \xv_t) \Big\vert \xv_t\right] \right\|^2 + \underset{D(m_0^k)}{\Eb} \left\|\Eb_{m^k_{0|t}}\left[\tfrac{1}{t}(\xv_0 - \xv_t) \Big\vert \xv_t\right] - \sv_t(\xv_t) \right\|^2 \\
	 & \hspace{5em} + \underset{D(m_0^k)}{\Eb} \inner{\tfrac{1}{t}(\xv_0 - \xv_t) - \Eb_{m^k_{0|t}}\big[\tfrac{1}{t}(\xv_0 - \xv_t) \big\vert \xv_t\big]}{\Eb_{m^k_{0|t}}\big[\tfrac{1}{t}(\xv_0 - \xv_t) \Big\vert \xv_t\big] - \sv_t(\xv_t)}. 
\end{align*}
For the last term,
\begin{align*}
	&\underset{D(m_0^k)}{\Eb} \inner{\tfrac{1}{t}(\xv_0 - \xv_t) - \Eb_{m^k_{0|t}}\big[\tfrac{1}{t}(\xv_0 - \xv_t) \big\vert \xv_t\big]}{\Eb_{m^k_{0|t}}\big[\tfrac{1}{t}(\xv_0 - \xv_t) \Big\vert \xv_t\big] - \sv_t(\xv_t)} \\
	=& \underset{m_t^k}{\Eb} \inner{\Eb_{m^k_{0|t}}\big[\tfrac{1}{t}(\xv_0 - \xv_t) \big\vert \xv_t\big] - \Eb_{m^k_{0|t}}\big[\tfrac{1}{t}(\xv_0 - \xv_t) \big\vert \xv_t\big]}{\Eb_{m^k_{0|t}}\big[\tfrac{1}{t}(\xv_0 - \xv_t) \Big\vert \xv_t\big] - \sv_t(\xv_t)} \\
	=& \underset{m_t^k}{\Eb} \inner{\zero}{\Eb_{m^k_{0|t}}\big[\tfrac{1}{t}(\xv_0 - \xv_t) \Big\vert \xv_t\big] - \sv_t(\xv_t)}  = 0. 
\end{align*}
As a result, 
\begin{align*}
	&\underset{D(m_0^k)}{\Eb} \left\|\tfrac{1}{t}(\xv_0 - \xv_t) - \sv_t(\xv_t) \right\|^2\\
	=& \underbrace{\underset{D(m_0^k)}{\Eb} \left\|\tfrac{1}{t}(\xv_0 - \xv_t) - \Eb_{m^k_{0|t}}\left[\tfrac{1}{t}(\xv_0 - \xv_t) \Big\vert \xv_t\right] \right\|^2}_{\text{Independent of $\sv$} \, \Rightarrow \, \text{Const.}} + \underset{D(m_0^k)}{\Eb} \left\|\Eb_{m^k_{0|t}}\left[\tfrac{1}{t}(\xv_0 - \xv_t) \Big\vert \xv_t\right] - \sv_t(\xv_t) \right\|^2. 
\end{align*}
Thus, 
\begin{align*}
	\argmin_\sv \, \Lc^\dagger(\sv, M^k) &= \argmin_\sv \underset{D(m_0^k)}{\Eb} \int_0^\tau \left\|\Eb_{m^k_{0|t}}\left[\frac{1}{t}(\xv_0 - \xv_t) \Big\vert \xv_t\right] - \sv_t(\xv_t) \right\|^2 \diff t \\
	&= \argmin_\sv ~~ \int_0^\tau \underset{D(m_0^k)}{\Eb} \left\|\tfrac{1}{t}(\xv_0 - \xv_t) - \sv_t(\xv_t) \right\|^2 \diff t + \text{Const.} \\
	&=  \argmin_\sv ~~ \int_0^\tau \underset{D(m_0^k)}{\Eb} \left\|\tfrac{1}{t}(\xv_0 - \xv_t) - \sv_t(\xv_t) \right\|^2 \diff t. 
\end{align*}

\subsection{Optimal Solution to \texorpdfstring{\cref{def:gamma_diff_proj}}{}}
\label{appx:opt_sol_gamma_diff_proj}

Note that, by the disintegration theorem (e.g., see \citealt{VargasTLL2021}, Appx B),
\begin{align*}
	&~\underset{P \in \Dc }{\argmin} ~ (1-\gamma)  \KL{ P^{k, \gamma} }{P} + \gamma \KL{ M^k }{P}\\
	=\, & ~~ \underset{P \in \Dc }{\argmin} \, (1-\gamma) \left[\KL{p_0^{k, \gamma}}{p_0} + \underbrace{\underset{P^{k, \gamma}}{\Eb}\Big[\log \frac{\diff P^{k, \gamma}(\cdot \vert \xv_0)}{\diff P(\cdot \vert \xv_0)}\Big]}_{\text{Const.}} \right] \\
	&\hspace{15em} + \gamma \left[ \KL{m_0^k}{p_0} + \underbrace{\underset{{M^k}}{\Eb} \Big[\log \frac{\diff M^k(\cdot \vert \xv_0)}{\diff P(\cdot \vert \xv_0)}\Big]}_{\text{Const.}} \right]\\
	=\,& ~~ \underset{P \in \Dc }{\argmin} \, (1-\gamma) \KL{p_0^{k, \gamma}}{p_0} + \gamma \KL{m_0^k}{p_0} \\
	=\, & ~~ \underset{P \in \Dc }{\argmin} \, -\int_{\Rb^d} \left[(1-\gamma) \, p_0^{k, \gamma}(\xv_0) + \gamma \, m_0^k(\xv_0) \right] \log p_0(\xv_0) \, \diff \xv_0 + \text{Const.} \\
	=\, & ~~ \underset{P \in \Dc }{\argmin} \,\, \KL{(1-\gamma)p_0^{k, \gamma} + \gamma m_0^k}{p_0}. 
\end{align*}
As a result,
\begin{align}
	p_0^{k+1, \gamma} = (1 - \gamma) \, p_0^{k, \gamma} + \gamma \, m_0^{k}. 
\end{align}

\subsection{Results Related to SFBD Flow}

\convGammaSFBD*
\begin{proof}
	Let $P^*$ denote the path measure induced by the forward process \cref{eq:fwd_diff} with $p_0 = p_\text{data}$. In addition, let $\Fc(q) = \KL{p_\text{data}}{q}$. For brevity, we drop the $\gamma$ in $P^{k,\gamma}$ and its marginal distributions~$p_0^{k, \gamma}$ and $p_\tau^{k, \gamma}$. 
	
	Note that,
	\begin{align}
		\KL{P^*}{M^k} = \Fc(m_0^k) + \underbrace{\Eb_{P^*}\left[ \frac{1}{2 } \int_0^\tau \|\bv^{k}(\xv_t, t)\|^2 \diff t \right]}_{:= \Bc_k}, \label{eq:convGammaSFBD:1}
	\end{align}
	where $\bv^{k}(\xv_t, t)$ is the drift of the forward process inducing $M^k$ with $\xv_0 \sim m_0^k$.
	
	In addition, through the convexity of the KL divergence, 
	\begin{align*}
		\Fc \big( p_0^{k+1} \big) = \Fc \big((1-\gamma) p_0^k + \gamma m_0^k \big) \leq (1-\gamma) \Fc(p_0^k) + \gamma \Fc(m_0^k),
	\end{align*}
	which implies,
	\begin{align}
		\Fc(m_0^k) \geq \Fc(p_0^k) + \tfrac{1}{\gamma} \big(\Fc(p_0^{k+1}) - \Fc(p_0^k)\big). \label{eq:convGammaSFBD:2}
	\end{align}
	
	As a result, 	
	\begin{align*}
		\Fc(p_0^k) &= \KL{P^*}{P^k} = \KL{p_\tau^*}{p_\tau^k} + \Eb_{p^*} \left[\int_0^\tau \frac{1}{2 } \big\| \nabla\log p_t(\xv_t) -  \sv_t^k(\xv_t)\big\|^2\right]\\
		&= \KL{p_\tau^*}{p_\tau^k} + \KL{P^*}{M^k} 
		\overset{\eqref{eq:convGammaSFBD:1}}{=} \KL{p_\tau^*}{p_\tau^k} + \Fc(m_0^k) + \Bc_k \\
		&\overset{\eqref{eq:convGammaSFBD:2}}{\geq} \KL{p_\tau^*}{p_\tau^k} +  \Bc_k  + \tfrac{1}{\gamma}\Big(\Fc(p_0^{k+1}) - \Fc(p_0^k)\Big) + \Fc(p_0^k)\\
		&\geq \KL{p_\tau^*}{p_\tau^k} +  \tfrac{1}{\gamma}\Big(\Fc(p_0^{k+1}) - \Fc(p_0^k)\Big) + \Fc(p_0^k). \numberthis \label{eq:convGammaSFBD:2.1}
	\end{align*}
	Rearrangement yields
	\begin{align}
        \KL{p_{\rm data}}{p_0^{k + 1, \gamma}} \hspace{-0.12em}  - \KL{p_{\rm data}}{p_0^{k, \gamma}} \leq -\gamma  \KL{p_\tau^*}{p_\tau^{k, \gamma}}, \label{eq:convGammaSFBD:3}
	\end{align}
	the monotonicity of $p_0^{k, \gamma}$ in $k$ in the proposition. Equivalently,
	\begin{align}
		 \Fc(p_0^{k + 1, \gamma}) -  \Fc(p_0^{k, \gamma}) \leq -\gamma  \KL{p_\tau^*}{p_\tau^{k, \gamma}}. 
	\end{align}

	
	Telescoping it yields:
	\begin{align}
		\Fc(p_0^{0, \gamma}) = \sum_{k=0}^K \Fc(p_0^{k, \gamma}) - \Fc(p_0^{k+1, \gamma}) \geq \gamma \sum_{k=1}^K \KL{p^*_\tau}{p^{k, \gamma}_\tau}. 
	\end{align}
	Thus,
	\begin{align}
		\min_{k \in \{1, 2, \ldots, K\}} \KL{p^*_\tau}{p^{k, \gamma}_\tau} \leq \frac{\Fc(p_0^{0, \gamma})}{\gamma K} = \frac{\Fc(p_{\Ec_\text{clean}})}{\gamma K}. 
	\end{align}
	Applying \cref{prop:conv_identify}, we get
	\begin{align}
			\min_{k \in \{1, 2, \ldots, K\}} \left|\Phi_{p_\text{data}}(\uv) - \Phi_{p_0^{k, \gamma}}(\uv)\right|\leq \exp \big(\frac{\tau}{2} \|\uv\|^2 \big) \left(\frac{2 \KL{p_{\rm data}}{p_{\Ec_\text{clean}}}}{\gamma K}\right)^{\sfrac12}.
	\end{align}
\end{proof}

\NonParaUpdateV*
\begin{proof}
	For $t \in [0, \tau]$, let $\phiv$ be a function of the same function space as $\sv_t^k$ and $p_0$ the density of a distribution defined on $\Rb^d$. Then for $\epsilon \in (0,1]$, we have
	\begin{align*}
		\Lc_t(\sv_t + \epsilon \phiv, p_0)
		=& \underset{D(p_0)}{\Eb} \left[\frac{1}{2} \Big\| \frac{\xv_0 - \xv_t}{t} - (\sv_t + \epsilon {\phiv}) (\xv_t) \Big\|^2\right] \\
		=& \Lc_t(\xv_t, p_0) + \epsilon\underset{D(p_0)}{\Eb} \left[\inner{\sv_t(\xv_t) - \frac{\xv_0-\xv_t}{t}}{ \phiv(\xv_t)}\right] + o(\epsilon) \\
		=& \Lc_t(\xv_t, p_0) + \epsilon \inner{\phiv(\xv_t)}{\Big(\sv_t(\xv_t) - \frac{\xv_0-\xv_t}{t}\Big) \, \diff P_{0t}(\xv_0, \xv_t)} \\
		=& \Lc_t(\xv_t, p_0) + \epsilon \inner{\phiv(\xv_t)}{\Big(\sv_t(\xv_t) - \frac{\xv_0-\xv_t}{t}\Big) \, \diff P_{0t} (\xv_0, \xv_t)} \\
		=& \Lc_t(\xv_t, p_0) + \epsilon \inner{\phiv(\xv_t)}{\Big(\sv_t(\xv_t) - \frac{\Eb_{P_{0|t}}[\xv_0|\xv_t] - \xv_t}{t}\Big) \,  d P_{t}(\xv_t)}. 
	\end{align*}
	As a result, 
	\begin{align}
		\nabla_\mu \Lc(\sv_t, p_0) (\xv_t) =  \Big(\sv_t(\xv_t) - \frac{\Eb_{P_{0|t}}[\xv_0|\xv_t]- \xv_t}{t} \Big) \frac{\diff P_t}{\diff \mu}(\xv_t). \label{eq:drift_func_grad_descent:proof:1}
	\end{align}

    We note that when $k = 0$, $\sv_t^0(\xv_t)$ is pretrained on $P_{\Ec_\text{clean}}$. As a result, by e.g., \citep[Lemma 1]{SongDCS2023}, 
    \begin{align}
		\sv_t^0(\xv_t) =  \frac{\Eb_{{ (P_{\Ec_\text{clean}}})_{0|t} }(\xv_0 \vert \xv_t) - \xv_t }{t} = \frac{\Eb_{P^{0, \gamma}_{0|t}}(\xv_0 \vert \xv_t)- \xv_t }{t}
	\end{align}    
    for any $t\in [0,\tau]$ and $\xv_t\in \Rb^d$, which is the negative backward drift of $M^{0}$ in $\gamma$-SFBD.

	Then assume that for $k \in \Nb$, for any $t\in [0,\tau]$ and $\xv_t\in \Rb^d$, we have 
	\begin{align}
		\sv_t^k(\xv_t) = \frac{\Eb_{P^{k, \gamma}_{0|t} }(\xv_0 \vert \xv_t)- \xv_t}{t}, \label{eq:drift_func_grad_descent:proof:1.5}
	\end{align}
	correpsonding to the negative backward drift of $M^{k}$ in $\gamma$-SFBD.	
	
	Then for $k+1$, \cref{eq:drift_func_grad_descent}  gives
	\begin{align*}
		\sv^{k+1}_t(\xv_t) &= \sv_t^k (\xv_t) - \gamma \,  \nabla_{P_t^{k+1, \gamma}} \Lc_t(\sv_t^k, m_0(\sv^k)) (\xv_t) \\
		&\overset{\eqref{eq:drift_func_grad_descent:proof:1}}{=} \sv_t^k(\xv_t) - \gamma \Big(\sv^k_t(\xv_t) - \frac{\Eb_{D(m_0^k)_{0|t}}[\xv_0|\xv_t]- \xv_t}{t} \Big) \frac{\diff D(m_0^k)_t}{\diff P_t^{k+1, \gamma}}(\xv_t) \\ 
		&= \big(1- \delta(\xv_t)\big) \, \sv_t^k(\xv_t) + \delta(\xv_t) \, \frac{ \Eb_{D(m_0^k)_{0|t}}[\xv_0|\xv_t] - \xv_t}{t},\numberthis\label{eq:drift_func_grad_descent:proof:2}
	\end{align*}
	where $\delta(\xv_t) = \gamma \frac{\diff D(m_0^k)_t}{\diff P_t^{k+1, \gamma}}(\xv_t) $. We note that, by \cref{eq:gamma_diff_proj_t=0},
	\begin{align}
		p_0^{k+1, \gamma} = (1 - \gamma) \, p_0^{k, \gamma} + \gamma \, m_0^{k}. 
	\end{align}
	As a result, 
	\begin{align}
		P_t^{k+1, \gamma} = (1 - \gamma) \, P_t^{k, \gamma} + \gamma \, D(m_0^{k})_t
	\end{align}
	and 
	\begin{align}
		\delta(\xv_t) =  \frac{\gamma\diff D(m_0^k)_t}{\diff (1 - \gamma) \, P_t^{k, \gamma} + \gamma \, D(m_0^{k})_t}(\xv_t), &~~ 1 - \delta(\xv_t) =  \frac{(1-\gamma)\diff P_t^{k, \gamma}}{\diff (1 - \gamma) \, P_t^{k, \gamma} + \gamma \, D(m_0^{k})_t}(\xv_t). 
	\end{align}
	Thus,
	\begin{align*}
		\sv_t^{k+1}(\xv_t) &\overset{\eqref{eq:drift_func_grad_descent:proof:2}}{=} \sv_t^k(\xv_t) \, \frac{(1-\gamma)\diff P_t^{k, \gamma}}{\diff (1 - \gamma) \, P_t^{k, \gamma} + \gamma \, D(m_0^{k})_t}(\xv_t)  \\
		& ~~~~~~~~~~~~~~~~~~ + \frac{\Eb_{D(m_0^k)_{0|t}}[\xv_0|\xv_t]- \xv_t}{t} \, \frac{\gamma\diff D(m_0^k)_t}{\diff (1 - \gamma) \, P_t^{k, \gamma} + \gamma \, D(m_0^{k})_t}(\xv_t) \\
		& \overset{\eqref{eq:drift_func_grad_descent:proof:1.5}}{=}
		\frac{\Eb_{P^{k, \gamma}_{0|t}}(\xv_0 \vert \xv_t) - \xv_t}{t} \, \frac{(1-\gamma)\diff P_t^{k, \gamma}}{\diff (1 - \gamma) \, P_t^{k, \gamma} + \gamma \, D(m_0^{k})_t}(\xv_t)  \\
		& ~~~~~~~~~~~~~~~~~~ + \frac{\Eb_{D(m_0^k)_{0|t}}[\xv_0|\xv_t] - \xv_t}{t} \, \frac{\gamma\diff D(m_0^k)_t}{\diff (1 - \gamma) \, P_t^{k, \gamma} + \gamma \, D(m_0^{k})_t}(\xv_t) \\
		&= - \frac{1}{t} \xv_t + \frac{1}{t} \int_{\xv'_0 \in \Rb^d} \xv'_0 \frac{\diff \, (1-\gamma)P_{0t}^{k,\gamma} + \gamma D(m_0^k)_{0t}}{\diff \,  (1 - \gamma) \, P_t^{k, \gamma} + \gamma \, D(m_0^{k})_t} (\xv'_0, \xv_t) \\
		&= - \frac{1}{t} \xv_t + \frac{1}{t} \int_{\xv'_0 \in \Rb^d} \xv'_0 \frac{\diff \, P_{0t}^{k+1, \gamma}}{\diff \,  P_t^{k+1, \gamma}} (\xv'_0, \xv_t)  \\
		&= \frac{\Eb_{P^{k+1, \gamma}_{0|t}}(\xv_0 \vert \xv_t) - \xv_t}{t},
	\end{align*}
	which is the negative backward drift of $M^{k+1}$. 
\end{proof}

\convCtnSFBD*
\begin{proof}
	According to \cref{eq:convGammaSFBD:3}, we have
	\begin{align}
        \frac{1}{\gamma} \left(\KL{p_{\rm data}}{p_0^{k + 1, \gamma}}  - \KL{p_{\rm data}}{p_0^{k, \gamma}}\right) \leq - \KL{p_\tau^*}{p_\tau^{k, \gamma}},  \label{eq:convCtnSFBD:1}
	\end{align}
	for all $\gamma > 0$ and $k \in \Nb$. 
	
	Fix $\kappa > 0$ and let $\{\gamma_i\} \to 0$ with $k_i = \kappa/\gamma_i \in \Nb$. Then $p_0^{k_i, \gamma_i} \to p_0^\kappa$ via Euler approximation. Taking the limit yields:
	\begin{align*}
		\frac{\diff}{\diff \kappa} \KL{p_{\rm data}}{p_0^\kappa} &= \lim_{i\rightarrow \infty} \frac{1}{\gamma_i} \left(\KL{p_{\rm data}}{p_0^{k_i+ 1, \gamma_i}}- \KL{p_{\rm data}}{p_0^{k_i , \gamma_i}} \right) \\
		&\overset{\eqref{eq:convCtnSFBD:1}}{\leq} \lim_{i \rightarrow \infty} - \KL{p^*_\tau}{p_\tau^{k_i, \gamma_i}} = - \KL{p^*_\tau}{p_\tau^{\kappa}}, \numberthis \label{eq:convCtnSFBD:2}
	\end{align*}
	establishing the monotonicity claim.
	
	In addition, integrating both sides of \cref{eq:convCtnSFBD:2} over $[0, \Kc]$ gives:
	\begin{align}
		\KL{p_\text{data}}{p_0^\Kc} - \KL{p_\text{data}}{p_0^0} \leq - \int_0^{\Kc}  \KL{p_\tau^*}{p_\tau^{\kappa}} \diff \kappa. 
	\end{align}
	As a result, 
	\begin{align*}
		\inf_{\kappa \in [0, \Kc]} \KL{p_\tau^*}{p_\tau^\kappa} \leq \frac{1}{\Kc} \KL{p_{\rm data}}{p_0^0} = \frac{1}{\Kc} \KL{p_{\rm data}}{p_{\Ec_\text{clean}}}. 
	\end{align*}
	Applying \cref{prop:conv_identify} concludes the convergence argument in the corollary.
\end{proof}

\convGammaSFBDInperfect*
\begin{proof}
The proof follows a similar idea used in the proof of \cref{prop:convGammaSFBD}. Recall \cref{eq:convGammaSFBD:2.1}:
\begin{align*}
		\Fc(p_0^k) &= \KL{P^*}{P^k} = \KL{p_\tau^*}{p_\tau^k} + \Eb_{p^*} \left[\int_0^\tau \frac{1}{2 } \big\| \nabla\log p_t(\xv_t) -  \sv_t^k(\xv_t)\big\|^2\right]\\
		&= \KL{p_\tau^*}{p_\tau^k} + \KL{P^*}{M^k} \geq \KL{p_\tau^*}{p_\tau^k} +  \tfrac{1}{\gamma}\Big(\Fc(p_0^{k+1}) - \Fc(p_0^k)\Big) + \Fc(p_0^k). 
\end{align*}
When $\sv^k$ is approximated by $\hat\sv^k$, and $\gamma = \gamma_k$, we instead have 
\begin{align*}
	\Fc(p_0^k)  &= \KL{p_\tau^*}{p_\tau^k} + \KL{P^*}{M^k} \geq \KL{p_\tau^*}{p_\tau^k} + \KL{P^*}{\hat M^k} - \delta_k \\
	&\geq \KL{p_\tau^*}{p_\tau^k} +  \tfrac{1}{\gamma_k}\Big(\Fc(p_0^{k+1}) - \Fc(p_0^k)\Big) + \Fc(p_0^k) - \delta_k.  
\end{align*}
Cancelling out $\Fc(p_0^k)$ on both sides followed by rearrangement yields:
\begin{align}
	\Fc(p_0^k) - \Fc(p_0^{k+1}) \geq \gamma_k \Big(\KL{p_\tau^*}{p_\tau^k}  - \delta_k \Big). 
\end{align}
Apply telescoping to obtain
\begin{align}
	\Fc(p_0) \geq  \sum_{k=0}^{K-1} \gamma_k \KL{p_\tau^*}{p_\tau^k} - \sum_{k=0}^{K-1} \gamma_k \delta_k. 
\end{align}
As a result,
\begin{align}
	\Fc(p_0) + \sum_{k=0}^{K-1} \gamma_k \delta_k \geq \sum_{k=0}^{K-1} \gamma_k \KL{p_\tau^*}{p_\tau^k} \geq \left(\sum_{k=1}^{K-1} \gamma_k \right) \left(\frac{1}{K}\sum_{k=1}^{K-1} \KL{p_\tau^\ast}{p_\tau^k}\right),
\end{align}
where the last inequality is by Chebyshev's sum inequality. As a result, 
\begin{align}
	\frac{1}{K} \sum_{k=0}^{K-1} \KL{p_\tau^*}{p_\tau^k}  \leq \frac{\Fc(p_0)}{\sum_{k=0}^{K-1} \gamma_k}  + \frac{\sum_{k=0}^{K-1} \gamma_k \delta_k}{\sum_{k=0}^{K-1} \gamma_k}. 
\end{align}
Assume we can show $\sum_k \gamma_k \rightarrow \infty$ and $\sum_k\gamma_k \delta_k < \infty$ (also known as the Robbins-Monro requirements). Then the left-hand side goes to zero as $K \rightarrow \infty$. Therefore, $\KL{p_\tau^*}{p_\tau^k} \rightarrow 0$. Then, applying the same argument as the one in the proof of \cref{prop:convGammaSFBD}, we conclude $p_0^k \rightarrow p_\text{data}$. 

We complete the proof by showing that the Robbins-Monro requirements are indeed satisfied. 

Recall that the step size at iteration $k$ is defined by
\begin{align}
    \gamma_k &= \min\!\left(\frac{\gamma_0}{\sqrt{v_k}},\, \Gamma\right), 
    &
    v_k &= \rho v_{k-1} + (1-\rho)\delta_k^2 ,
\end{align}
with $\rho,\Gamma \in (0,1)$, $\gamma_0,v_0>0$, and $\sum_{k=1}^\infty \delta_k < \infty$.

\medskip
\noindent\textbf{Decay of $v_k$.}
Unrolling the recursion gives
\begin{align}
    v_k = \rho^k v_0 + (1-\rho)\sum_{j=1}^k \rho^{\,k-j}\,\delta_j^2 .
\end{align}
Since $\rho\in(0,1)$, the first term vanishes as $k\to\infty$.  
Moreover, $\sum_j \delta_j^2 < \infty$ (because $\delta_j^2 \le \delta_j$ eventually), so the convolution with the geometric kernel also vanishes.  
Therefore,
\(
v_k \to 0 \quad \text{as } k\to\infty
\). 

\medskip
\noindent\textbf{Divergence of $\sum_k \gamma_k$.}
Because $v_k \to 0$, for large $k$ we have $\gamma_0/\sqrt{v_k} \ge \Gamma$, hence
\[
\gamma_k = \Gamma > 0 \quad \text{eventually}.
\]
Therefore
\[
\sum_{k=1}^\infty \gamma_k \;\ge\; \sum_{k=K}^\infty \Gamma \;=\; \infty,
\]
so the first Robbins--Monro condition holds.

\medskip
\noindent\textbf{Convergence of $\sum_k \gamma_k \delta_k$.}
For all $k$,
\[
\gamma_k \le \Gamma \quad\implies\quad \gamma_k\delta_k \le \Gamma \delta_k .
\]
Hence
\[
\sum_{k=1}^\infty \gamma_k \delta_k 
\;\le\; \Gamma \sum_{k=1}^\infty \delta_k 
\;<\;\infty,
\]
by assumption.  

\medskip
\noindent
Thus both Robbins--Monro requirements are satisfied:
\[
\sum_{k=1}^\infty \gamma_k = \infty,
\qquad
\sum_{k=1}^\infty \gamma_k \delta_k < \infty.
\]
\end{proof}

\subsection{Construction of $\delta_k$ for the $\gamma$-Adaptive Online SFBD Algorithm}
\label{appx:delta_k_geometric_decay}
In this section, we provide additional implementation details on how we construct $\delta_k$ for the $\gamma$-adaptive Online SFBD algorithm.  

As indicated in \cref{prop:convGammaSFBDInperfect}, we need to design a sequence $\{\delta_k\}$ satisfying \cref{eq:def_deltak}. Since the scaling factor can be absorbed into $\gamma_0$, and \cref{prop:convGammaSFBDInperfect} imposes no constraint on $\gamma_0$ other than positivity, our goal reduces to constructing $\delta_k$ such that
\begin{align}
	\delta_k \geq K \, \max\!\Big\{0,~\KL{P^\ast}{\hat M_k} - \KL{P^\ast}{M_k}\Big\}
\end{align}
for some $K>0$. Hence, we focus on constructing a sequence that reflects the relative variation in the difference between $\KL{P^\ast}{\hat M_k}$ and $\KL{P^\ast}{M_k}$.

Recall that, by \cref{appx:lem:kl_two_path}, we have
\begin{align*}
	\KL{P^\ast}{\hat M_k} = \KL{p_\tau^*}{p_\tau^*} + \Eb_{p^*} \left[\int_0^\tau \frac{1}{2 } \big\| \nabla\log p_t(\xv_t) -  \hat\sv_t^k(\xv_t)\big\|^2\right] = \Eb_{p^*}\left[\int_0^\tau \frac{1}{2} \big\| \nabla\log p_t(\xv_t) -  \hat\sv_t^k(\xv_t)\big\|^2 \right]. 
\end{align*}
In practice, this quantity is not directly accessible. Conventional diffusion models instead learn to align the estimated score $\hat\sv_t^k$ with the true score $\nabla \log p_t$ via conditional score matching, which relies on noisy estimates of the score function. However, this approach cannot be applied here, as it introduces excessive variance and requires direct access to the ground-truth clean data distribution $p_{\text{data}}$, which is unavailable in our setting.

Due to these challenges and the limited availability of clean samples, we instead measure the distance between the denoised samples generated using $\hat\sv_t^k$ and the clean samples $\Ec_\text{clean}$ in a suitable feature space as a proxy for the above quantity. This approximation is justified, as the quality of denoised samples directly depends on the accuracy of $\hat\sv_t^k$ across all $t \in [0, \tau]$.

While the Fréchet Inception Distance (FID)~\citep{HeuselRUNH2017} is the classical choice for quantifying such distances, its standard implementation typically requires at least 2,048 clean samples to ensure a well-conditioned covariance estimate -- substantially more than what is available in our setting. Moreover, FID is known to be biased when computed on small sample sets, which further limits its reliability under our data constraints. 

To address these issues, we adopt the Kernel Inception Distance (KID)~\citep{BinkowskiSAG2018}, which provides an unbiased estimator of the squared Maximum Mean Discrepancy (MMD) between feature embeddings and exhibits greater robustness with limited sample sizes. Empirically, we observe that the estimation remains stable and reliable even when the clean sample set contains as few as 50 images. 

Likewise, we measure the KID between the clean sample set and the running denoised sample set $\Ec$ as a practical proxy for $\KL{P^\ast}{M_k}$. Based on this, we define
\begin{align}
	\delta^2 \propto \max\Big(0,~\text{\sffamily KID}(\Ec_\text{clean}, \Ec_{\phiv}) - \text{\sffamily KID}(\Ec_\text{clean}, \Ec)\Big),
\end{align}
where we use $\delta^2$ instead of $\delta$ so that $v_k$ admits a clear interpretation as the exponential moving average of the KID difference $\delta_k^2$. This formulation makes $v_k$ a smoother and lower-variance estimator of the KID gap across iterations.





\textbf{Ensuring the Convergence of $\sum_k \delta_k$ via Skipping Denoised Set Updates}.
As discussed in \cref{sec:online_SFBD}, by conditionally skipping the denoised sample updates (i.e., setting $\lambda = 0$) as specified in \cref{alg:gamma_function}, we ensure that the effective sequence ${\delta_k}$ satisfies the convergence condition $\sum_k \delta_k < \infty$.

To see this, note that the alternative projection updates are performed only when the denoised sample set is refreshed. In other words, the deviation term $\delta_k$ of an iteration is counted as effective only when $\lambda > 0$ in that iteration. As a result, by \cref{alg:gamma_function}, we have
\[
v_k=\rho v_{k-1}+(1-\rho)\,\delta_k^2,\qquad 0<\rho<1,
\]
and the skipping mechanism ensures
\[
\delta_k^2\le \eta\, v_{k-1}
\]
for $\eta \in (0,1)$. Then
\[
v_k \le \bigl[\rho+(1-\rho)\eta\bigr]\; v_{k-1} =: \beta\, v_{k-1}.
\]
As 
$\eta<1$ (so $\beta= \rho+(1-\rho)\eta <1$),
\[
v_k \le \beta^k v_0 \quad\text{(geometric decay).}
\]
Plugging back into the $\delta_k^2$ bound gives
\[
\;\delta_k^2 \le \eta\, v_{k-1} \le \eta\,\beta^{\,k-1} v_0,\qquad
\beta=\rho+(1-\rho)\eta<1.
\]
So $\delta_k$ is upper-bounded by the geometric sequence 
$\sqrt{\eta v_0/\beta}\,\beta^{\frac{k}{2}}$. As a result, $\sum_k \delta_k  < \sqrt{\frac{\eta v_0}{\beta}} \sum_k \beta^{\frac{k}{2}} < \infty$.

Finally, we note that, in theory, $\hat{M}^k \rightarrow M^k$ given sufficiently many gradient descent updates. Consequently, $\gamma$ should not be set to zero infinitely often. In practice, however, due to the inherent approximation error of neural networks, the denoised set update naturally stabilizes and ceases near the end of the Online SFBD training.

\subsection{A Variant of \texorpdfstring{$\gamma$}{}-SFBD}
\label{appx:gamma_SFBD_v2}

Since the copyright-free clean samples are drawn from the true data distribution, it is practical to mix them with the denoised samples during denoiser updates to enhance overall sample quality. In particular, we generally believe that 
\begin{align}
	\Lc_\text{vis}(\alpha \, p_\text{clean} + (1-\alpha) \, p_\text{denoise}) \leq \Lc_\text{vis}(p_\text{denoise}),
	\label{eq:mix_clean_denoise}
\end{align}
where $\Lc_\text{vis}(p)$ denotes an aggregate loss that measures the visual quality of samples drawn from distribution $p$, and $p_\text{clean}$ and $p_\text{denoise}$ represent the distributions of clean and denoised samples, respectively. $\alpha$ depends on the ratios between the numbers of clean samples and the denoised samples. In practice, we have observed that this is always true when $\Lc_\text{vis}$ is instantiated as the FID.

We note that this heuristic technique can be naturally covered in our framework with little work. In particular, we can replace the original $\gamma$-Diff Proj with
\begin{flalign*}
    &\text{($\gamma$-Diff Proj-v2)} \quad 
    P^{k+1, \gamma} 
    = \argmin_{P \in \Dc} \; 
    \alpha\, \KL{P_{\text{clean}}}{P} 
    + (1 - \alpha) \left[
        (1 - \gamma)\, \KL{P^{k, \gamma}}{P} 
        + \gamma\, \KL{M^k}{P}
    \right]&
\end{flalign*}
where $P_\text{clean} = D(p_\text{clean})$ is a fixed diffusion process in $\Dc$ with the initial distribution having density $p_\text{clean}$. 

Applying a derivation similar to the one in \cref{appx:opt_sol_gamma_diff_proj}, again through the disintegration theorem, we have 
\begin{align*}
	&\underset{P \in \Dc}{\arg\min}~ \alpha \, \KL{P_\text{clean}}{P} + (1-\alpha) \Big[
        (1-\gamma) \, \KL{P^{k, \gamma}}{P} + \gamma \, \KL{M^k}{P} \Big]\\
        =& \underset{P \in \Dc}{\arg\min}~ \alpha \, \KL{p_\text{clean}}{p_0} + (1-\alpha) \Big[
        (1-\gamma) \, \KL{p_0^{k, \gamma}}{p_0} + \gamma \, \KL{m_0^k}{p_0}\Big] + \text{Const.} \\
        =& \, \underset{P \in \Dc }{\argmin} \, -\int_{\Rb^d} \alpha p_\text{clean}(\xv_0) + (1-\alpha)\left[(1-\gamma) \, p_0^{k, \gamma}(\xv_0) + \gamma \, m_0^k(\xv_0) \right] \log p_0(\xv_0) \, \diff \xv_0 + \text{Const.} \\
        =& \underset{P \in \Dc}{\arg\min}~ \KL{\alpha p_\text{clean} + (1-\alpha)\big[(1-\gamma) \, p_0^{k, \gamma} + \gamma \, m_0^k \big]}{p_0}. 
\end{align*}
As a result,
\begin{align*}
	\tilde p_0^{k+1, \gamma} &= \alpha \, p_\text{clean} + (1-\alpha) 
\left[(1 - \gamma) \, p_0^{k, \gamma} + \gamma \, m_0^{k}\right]  \\
&= \alpha \, p_\text{clean} + (1-\alpha) \, p_0^{k+1, \gamma},
\end{align*}
where \( p_0^{k+1, \gamma} \) is obtained via the standard $\gamma$-D-Proj defined in \cref{def:gamma_diff_proj}, and corresponds to \( p_\text{denoise} \) in \cref{eq:mix_clean_denoise}.

This variant of $\gamma$-D-Proj therefore recovers the exact update rule underlying the heuristic practice of mixing clean and denoised samples prior to fine-tuning the diffusion model.

Notably, when \( \gamma = 1 \), this variant reduces to a form of the original SFBD algorithm, which was heuristically employed in the initial SFBD paper~\citep{LuWY2025} to boost model performance—despite lacking theoretical justification at the time.

\subsection{Relationship to Consistency Constraint-based Methods}
\label{appx:rel_constency_based_mtd}
In \cref{sec:online_SFBD}, we argued that consistency-constraint-based (CC-based) methods enforcing consistency only between $r = 0$ and a positive time $s$ can be viewed as a special case of Online SFBD with $m = 1$ with $p^{k, \gamma}$ approximated by $m_0^k$. In this section, we elaborate on this connection and extend the discussion to more general CC-based methods that enforce consistency between arbitrary time steps $r < s$.

\textbf{Enforcing Consistency Between $r = 0$ and $s > 0$.} We assume the denoiser network satisfies $D_{\phiv}(\cdot, 0) = \textrm{Id}(\cdot)$, a condition explicitly enforced in EDM-based implementations. This design is both natural and intuitive, as $D_{\phiv}(\xv_0, 0)$ approximates $\Eb_{p_{0|t}}[\xv_0 \mid \xv_0]$ at $t = 0$, which is $\xv_0$ for any $p_{0t}$ induced by some distribution $p_0$ argumented by the forward transition kernel $p_{t|0}$ in \cref{def:diff_trans_kernel}. It has been adopted in all CC-based methods~\citep{DarasDD2024, DarasCD25}, the SFBD framework~\citep{LuWY2025}, and our work. 

\citet{LuWY2025} showed that, under this assumption, the denoising loss in \cref{eq:denoiser_loss} is equivalent to the consistency loss \cref{eq:consistency_loss}. For completeness, we include their derivation as follows:
\begin{align*}
	 &\underset{p_0}{\Eb} \,  \underset{p_{s|0}}{\Eb} \left[\|D_{\phiv}(\xv_s, s) - \xv_0 \|^2 \right] = \Eb_{p_s} \Eb_{p_{0|s}}\big[ \|D_{\phiv}(\xv_s, s) - \xv_0 \|^2  \big] \\
	 =\, & \Eb_{p_s} \Eb_{p_{0|s}}\big[ \|D_{\phiv}(\xv_s, s) - \Eb_{p_{0|s}}[\xv_0| \xv_s] + \Eb_{p_{0|s}}[\xv_0| \xv_s] - \xv_0 \|^2  \big] \\
	 =\, & \Eb_{p_s} \Eb_{p_{0|s}}\big[\|D_{\phiv}(\xv_s, s) - \Eb_{p_{0|s}}[\xv_0| \xv_s]  \|^2 \big] + \underbrace{\Eb_{p_s} \Eb_{p_{0|s}}\big[\|\Eb_{p_{0|s}}[\xv_0| \xv_s]  - \xv_0\|^2 \big]}_{\text{Const.}}  \\
	 &\hspace{10em} + 2 \underbrace{\Eb_{p_s} \Eb_{p_{0|s}}\big[\inner{D_{\phiv}(\xv_s, s) - \Eb_{p_{0|s}}[\xv_0| \xv_s]}{\Eb_{p_{0|s}}[\xv_0| \xv_s] - \xv_0} \big]}_{=0} \\ \numberthis \label{eq:rel_constency_based_mtd:1}
	 =\, & \Eb_{p_s}\big[\|D_{\phiv}(\xv_s, s) - \Eb_{p_{0|s}}[\xv_0| \xv_s]  \|^2 \big] + \textrm{Const.} \\  
	 \overset{(\text{arch ass})}{=} \hspace{-1em}&\, \hspace{1.5em} \Eb_{p_s} \big[\|D_{\phiv}(\xv_s, s) - \Eb_{p_{0|s}}[D_{\phiv}(\xv_0, 0)]  \|^2 \big] + \textrm{Const.},
	 \end{align*}
	 	 which is the consistency loss in \cref{eq:consistency_loss} when $r = 0$.
Therefore,
\begin{align}
\underset{p_0}{\Eb} \,  \underset{p_{s|0}}{\Eb} \left[\|D_{\phiv}(\xv_s, s) - \xv_0 \|^2 \right] \equiv \Eb_{p_s}\left[\|D_{\phiv}(\xv_s, s) - \Eb_{p_{0|s}}[D_{\phiv}(\xv_0, 0)]\|^2\right]\label{eq:denoise_consistency_eqv}
\end{align}
up to a constant, establishing the equivalence between the denoising loss used in \cref{alg:Online_SFBD} (M-proj) and the consistency loss in CC-based methods.

For Online SFBD, at the $k$-th iteration, we have 
\begin{align}
	p_0 = p_0^{k+1, \gamma} = (1 - \gamma)\, p_0^{k, \gamma} + \gamma\, m_0^k, 	
\end{align}
as presented in \cref{eq:gamma_diff_proj_t=0}, and $p_s$ is
\begin{align}
	p_s = p_0 \conv \Nc(0, s \Iv) = p_0^{k, \gamma}  \conv \Nc(0, s \Iv). 
\end{align}
Instead, in CC-based methods, 
\begin{align}
	p_0 \approx m_0^k. 
\end{align}

To see this, note that practical implementations of CC-based methods typically rely on two approximations:
\begin{itemize}
\item[(a)] $p_s$ is approximated using samples generated by adding Gaussian noise to corrupted data, where $s$ is chosen no less than the corruption level $\tau$~\citep{DarasCD25};
\item[(b)] $p_{0|s}$ is estimated via the backward SDE~\cref{eq:anderson_bwd}, with the drift term approximated by the current network (i.e., $\sv^k$).
\end{itemize}

For simplicity, we restrict the discussion to the case $s = \tau$. For the cases $s > \tau$, they reduce to the case $s = \tau$ under the assumption that the score function above $\tau$ is accurately estimated, which can be achieved by training the model through the ASM loss combined with the noisy samples \citep{DarasDD2024, DarasCD25}. These approximations essentially define the backward SDE process $M^k$, whose marginal at $t = 0$ is $m_0^k$, serving as the effective $p_0$ in CC-based training.

Note that CC-based methods form $(\xv_0, \xv_s)$ pairs using the backward SDE, whereas Online SFBD uses the forward process. As CC-based methods assume that corrupted samples can be approximated as drawn from $p_s$, the two pairing schemes are equivalent: both the forward and backward SDE yield the same joint distribution $p_{0s}(\xv_0, \xv_s)$, as discussed following \cref{eq:anderson_bwd}.

This approximation is reasonable when $m_0^k$ evolves slowly and $\gamma$ is not too small, as discussed in the main text. This condition typically holds in practice, as CC-based methods only take one gradient step per iteration. Moreover, CC-based methods often adopt the same pretraining strategy as OSFBD, allowing the network to learn global structure early on. As a result, drift updates during subsequent training are small, and $m_0^k$ changes slightly across iterations.

\textbf{Enforcing Consistency Between $r < s$.}
For any pair $r < s$, we note that 
\begin{align*}
	& \Eb_{p_s} \big[\|D_{\phiv}(\xv_s, s) - \Eb_{p_{0|s}}[D_{\phiv}(\xv_0, 0)]  \|^2 \big] \\
	\overset{(\text{arch ass})}{=} \hspace{-1em}&\, \hspace{1.5em} \Eb_{p_s} \left[\|D_{\phiv}(\xv_s, s) - \Eb_{p_{0|s}}[\xv_0 | \xv_s]\|^2 \right] \\
	=\, & \Eb_{p_s} \left[\left\|D_{\phiv}(\xv_s, s) - \Eb_{p_{r|s}} \left[\Eb_{p_{0|r}}[\xv_0 | \xv_r] \big| \xv_s \right]\right\|^2 \right] , \numberthis \label{eq:rel_constency_based_mtd:2}
\end{align*}
where the final equality uses the fact that the backward process is Markovian. In more detail, since the forward process is Brownian and thus Markovian, its time reversal (the backward process described by \cref{eq:anderson_bwd}) is also Markovian. Consequently, we can justify:
\begin{align*}
	\Eb_{p_{r|s}} \left[ \Eb_{p_{0|r}} [\xv_0 | \xv_r] \big| \xv_s \right] 
	&= \int \xv_0 \left( \int p_{0|r}(\xv_0 | \xv_r) \, p_{r|s}(\xv_r | \xv_s) \, \diff \xv_r \right) \diff \xv_0 \\
	&= \int \xv_0 \left( \int p_{0r|s}(\xv_0, \xv_r | \xv_s) \, \diff \xv_r \right) \diff \xv_0 \\
	&= \int \xv_0 \, p_{0|s}(\xv_0 | \xv_s) \, \diff \xv_0 \\
	&= \Eb_{p_{0|s}}[\xv_0 | \xv_s].
\end{align*}

As a result, by \cref{eq:rel_constency_based_mtd:2}, we have
\begin{align*}
	\Lc_\text{con}({\phiv}, 0, s) &= \Eb_{p_s} \big[\|D_{\phiv}(\xv_s, s) - \Eb_{p_{0|s}}[D_{\phiv}(\xv_0, 0)]  \|^2 \big] \\
	&=  \Eb_{p_s} \left[\left\|D_{\phiv}(\xv_s, s) - \Eb_{p_{r|s}} \left[\Eb_{p_{0|r}}[\xv_0 | \xv_r] \big| \xv_s \right]\right\|^2 \right] \\ 
	&\approx  \Eb_{p_s} \left[\left\|D_{\phiv}(\xv_s, s) - \Eb_{p_{r|s}} \left[D_{\textsf{stopgrad}(\phiv)}(\xv_r, r) \big| \xv_s \right]\right\|^2 \right] \\
	&= \Lc_\text{con}({\phiv}, r, s)
\end{align*}
where $\Eb_{p_{0|r}}[\xv_0 | \xv_r]$ is approximated using the current network, and \textsf{stopgrad} indicates a stop-gradient operation.

This suggests that enforcing consistency between arbitrary time pairs $r < s$ is effectively equivalent to enforcing it between $0$ and $s$, so the same argument for $r = 0$ applies.

\subsection{Auxiliary Results for References}
\label{appx:aux_results}

\begin{proposition}[{\citealt{LuWY2025}, Prop 1}]
\label{prop:conv_identify}
	Let $p$ and $q$ be two distributions defined on $\Rb^d$. For all $\uv \in \Rb^d$, 
	\begin{align*}
			|\Phi_p(\uv) -  \Phi_q(\uv)| \leq \exp \big(\frac{\tau}{2} \|\uv\|^2 \big) \sqrt{2 \,D_{\mathrm{KL}}(p \conv h\|q\conv h)},
	\end{align*}
	where $h \sim \Nc(\zero, \tau \Iv)$. 
\end{proposition}

\begin{lemma}[\citealt{PavonA1991}, \citealt{VargasTLL2021}]
\label{appx:lem:kl_two_path}
	Given two SDEs:
	\begin{align}
		\diff \xv_t = \fv_i(\xv_t, t) \diff t + \delta \diff \wv_t,~~~\xv_0\sim p_0^{(i)}(\xv)~~~~ t \in [0,T]
	\end{align}
	for $i = 1, 2$. Let $P^{(i)}_{0:T}$, for $i = 1, 2$, be the path measure induced by them, respectively. Then we have,
	\begin{align}
		\KL{P^{(1)}_{0:T}}{P^{(2)}_{0:T}} = \KL{p_0^{(1)}}{p_0^{(2)}} \; + \; \Eb_{P^{(1)}_{0:T}}\Bigl[\int_0^T \frac{1}{2}\,\|\fv_1(\xv_t,t)-\fv_2(\xv_t,t)\|^2\,dt\Bigr]. 
	\end{align}
	A similar result also applies to a pair of backward SDEs as well, where $p_0^{(i)}$ is replaced with~$p_T^{(i)}$. 
\end{lemma}
\begin{proof}
	By the disintegration theorem~(e.g., see \citealt[Appx B]{VargasTLL2021}), we have 
	 \begin{align}
	 	\KL{P_1}{P_2} = \KL{p_0^{(1)}}{p_0^{(2)}} \; + \; \Eb_{P^{(1)}_{0:T}}\left[\log \frac{\diff P^{(1)}_{0:T}(\cdot | \xv_0))}{\diff P^{(2)}_{0:T}(\cdot | \xv_0)} \right],
	 \end{align}
	 where $P^{(i)}_{0:T}(\cdot |\xv_0)$ is the conditioned path measure of $P^{(i)}_{0:T}$ given the initial point $\xv_0$. Then, applying the Girsanov theorem \citep{Kailath1971, Oksendal2003} on the second term yields the desired result. 
\end{proof}


\newpage

\section{SAMPLING RESULTS}
\label{appx:sample_results}
In this section, we present model-generated samples used to compute FID scores in \cref{sec:emp}, corresponding to the benchmark results in \cref{tb:performance_compare} and \cref{tb:performance_compare_addition}. We also include denoised samples at the final training step.

\subsection{CIFAR-10 (fixed $\gamma$)}
\begin{figure}[H]
    \centering
    \subfloat[Generated (FID: 3.22)]{
        \includegraphics[width=0.48\textwidth]{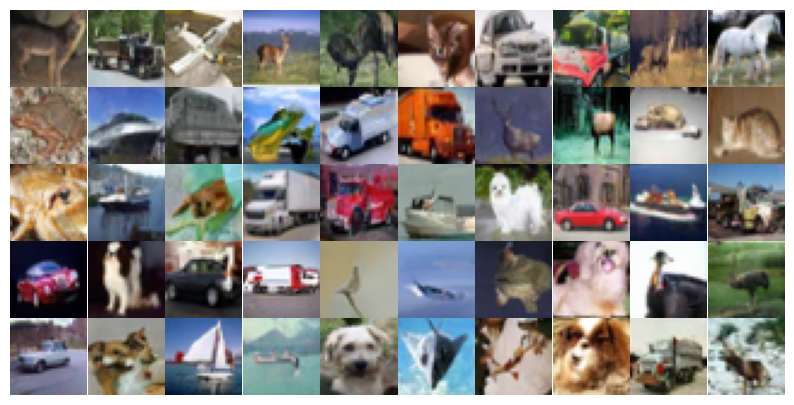}
    }
    \subfloat[Denoised (FID: 1.11)]{
        \includegraphics[width=0.48\textwidth]{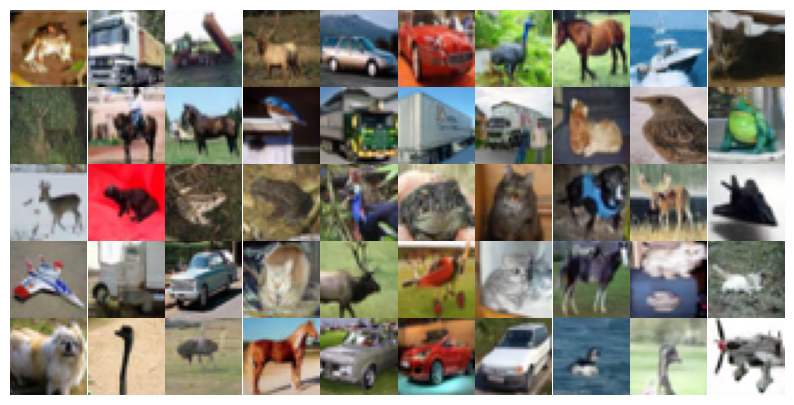}
    }
    \caption{50 clean samples, noise level $\sigma = 0.2$}
\end{figure}

\begin{figure}[H]
    \centering
    \subfloat[Generated (FID: 2.73)]{
        \includegraphics[width=0.48\textwidth]{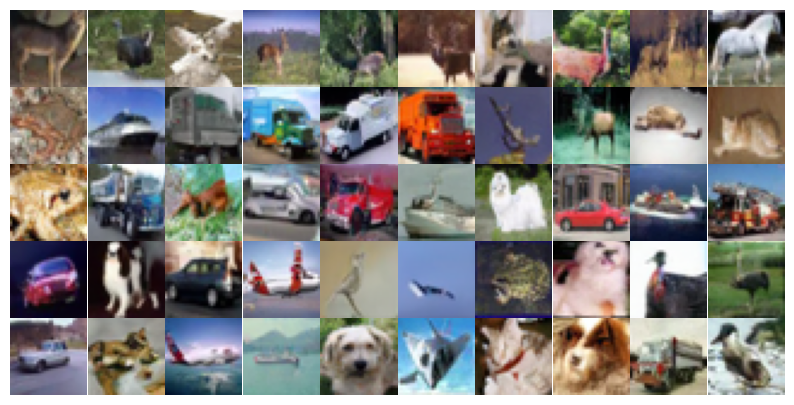}
    }
    \subfloat[Denoised (FID: 1.02)]{
        \includegraphics[width=0.48\textwidth]{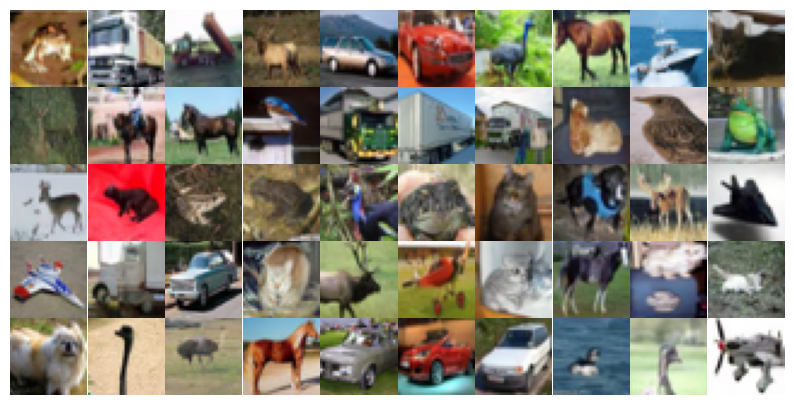}
    }
    \caption{5,000 clean samples (10\%), noise level $\sigma = 0.2$.}
\end{figure}

\begin{figure}[H]
    \centering
    \subfloat[Generated (FID: 6.56)]{
        \includegraphics[width=0.48\textwidth]{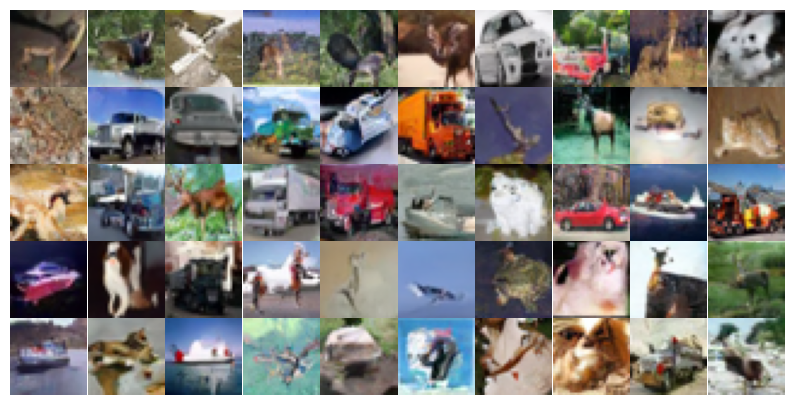}
    }
    \subfloat[Denoised (FID: 4.84)]{
        \includegraphics[width=0.48\textwidth]{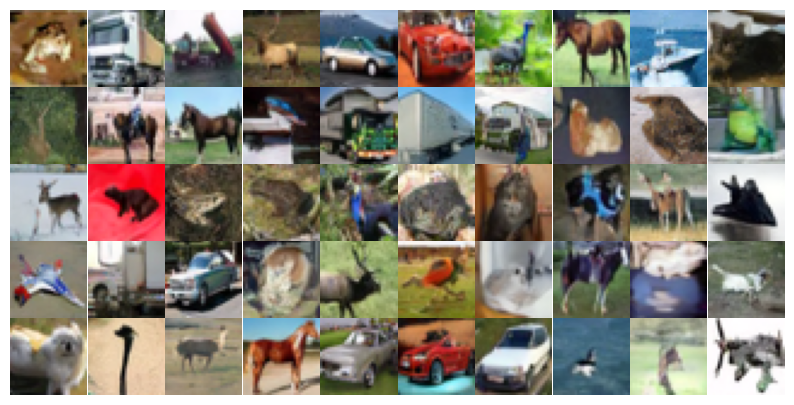}
    } 
    \caption{2,000 clean samples (4\%), noise level $\sigma = 0.59$.}
\end{figure}

\subsection{CIFAR-10 (adaptive $\gamma$)}
\begin{figure}[H]
    \centering
    \subfloat[Generated (FID: 3.12)]{
        \includegraphics[width=0.48\textwidth]{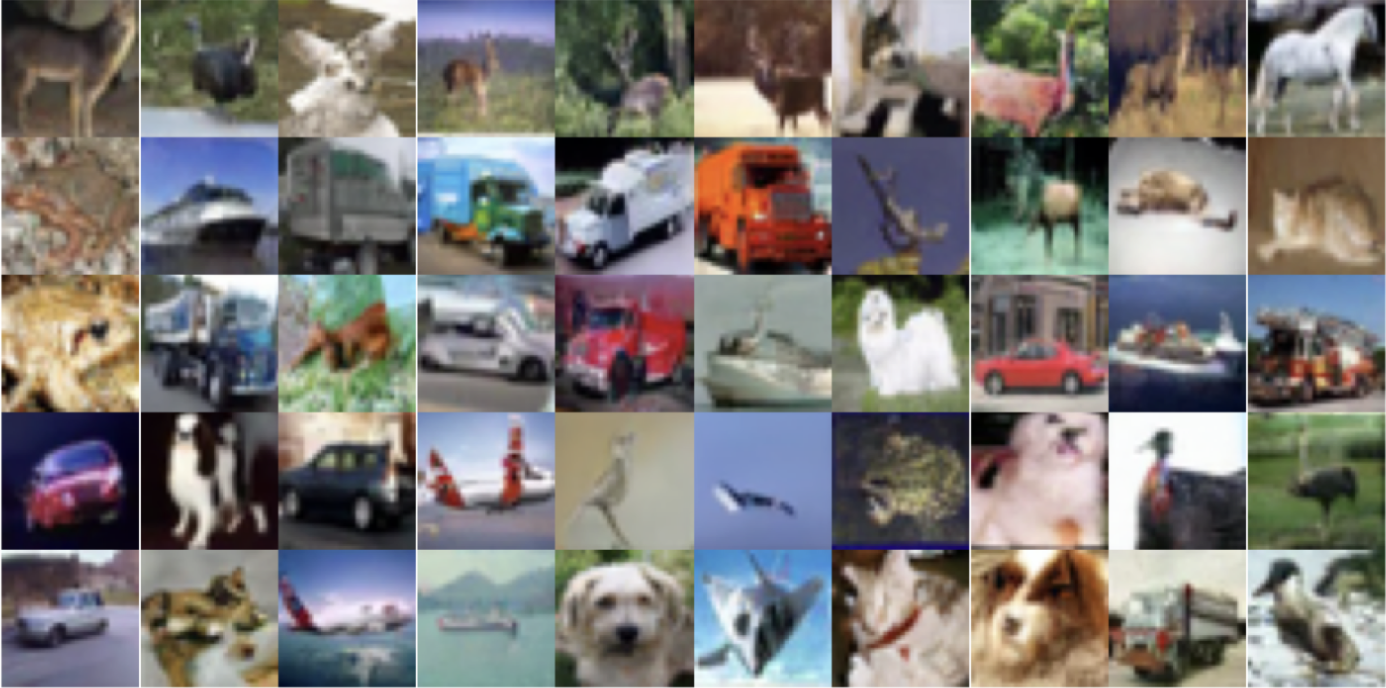}
    }
    \subfloat[Denoised (FID: 1.05)]{
        \includegraphics[width=0.48\textwidth]{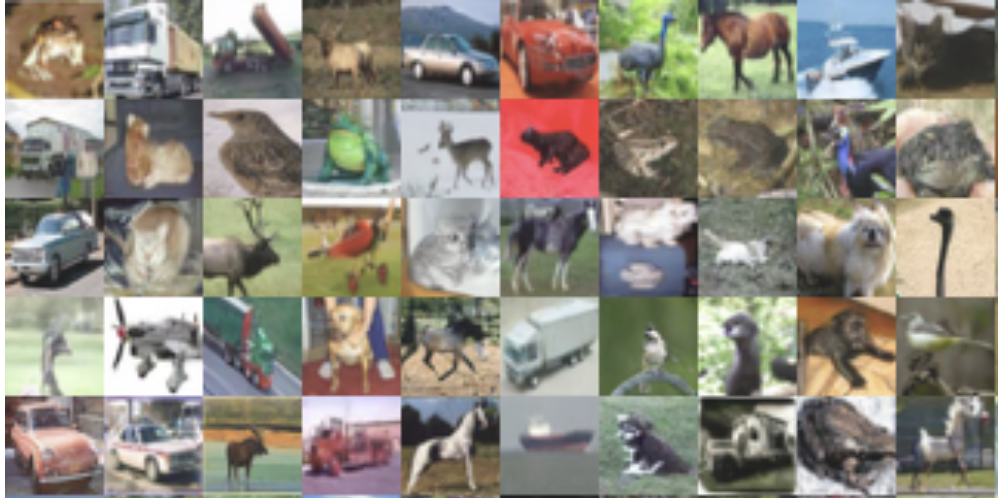}
    }
    \caption{50 clean samples, noise level $\sigma = 0.2$}
\end{figure}

\begin{figure}[H]
    \centering
    \subfloat[Generated (FID: 2.49)]{
        \includegraphics[width=0.48\textwidth]{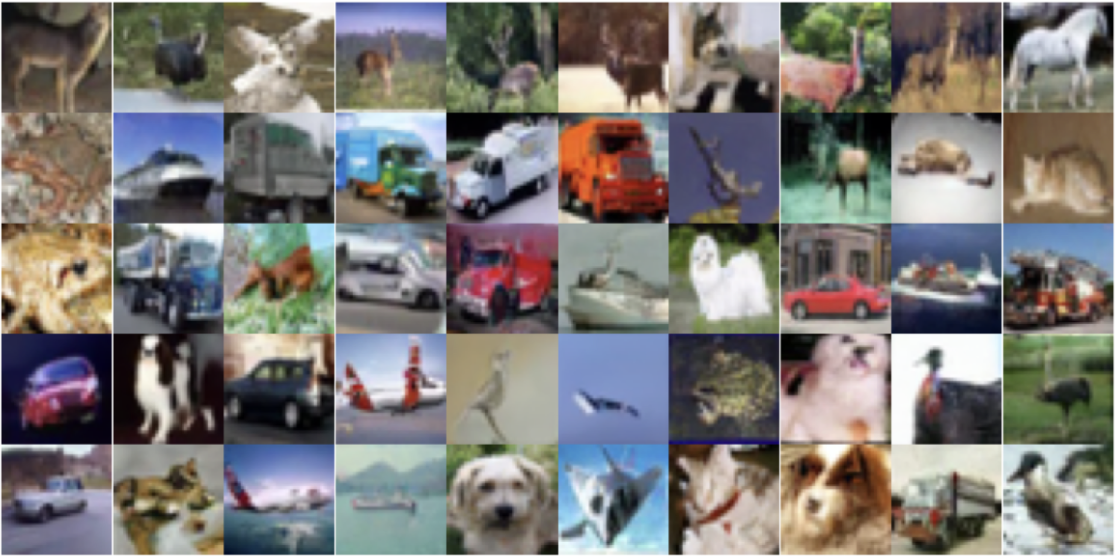}
    }
    \subfloat[Denoised (FID: 1.01)]{
        \includegraphics[width=0.48\textwidth]{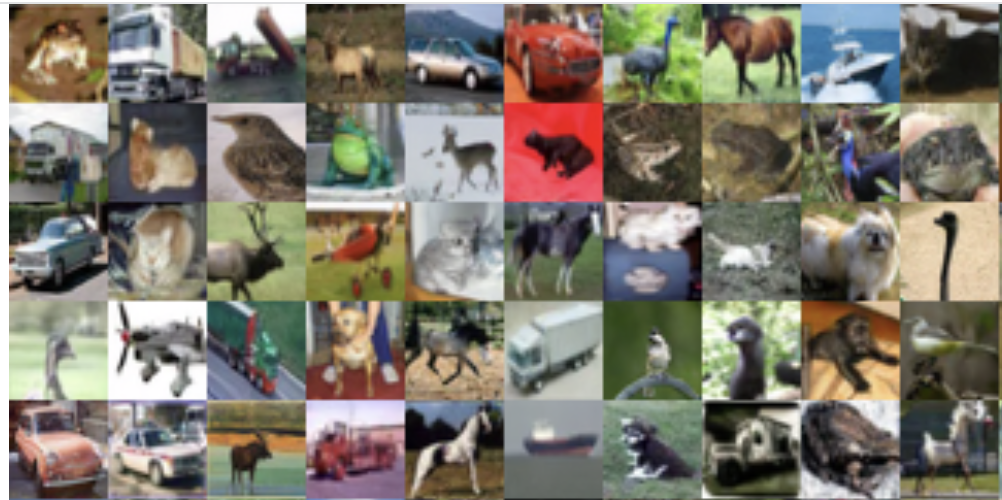}
    }
    \caption{5,000 clean samples (10\%), noise level $\sigma = 0.2$.}
\end{figure}

\begin{figure}[H]
    \centering
    \subfloat[Generated (FID: 6.21)]{
        \includegraphics[width=0.48\textwidth]{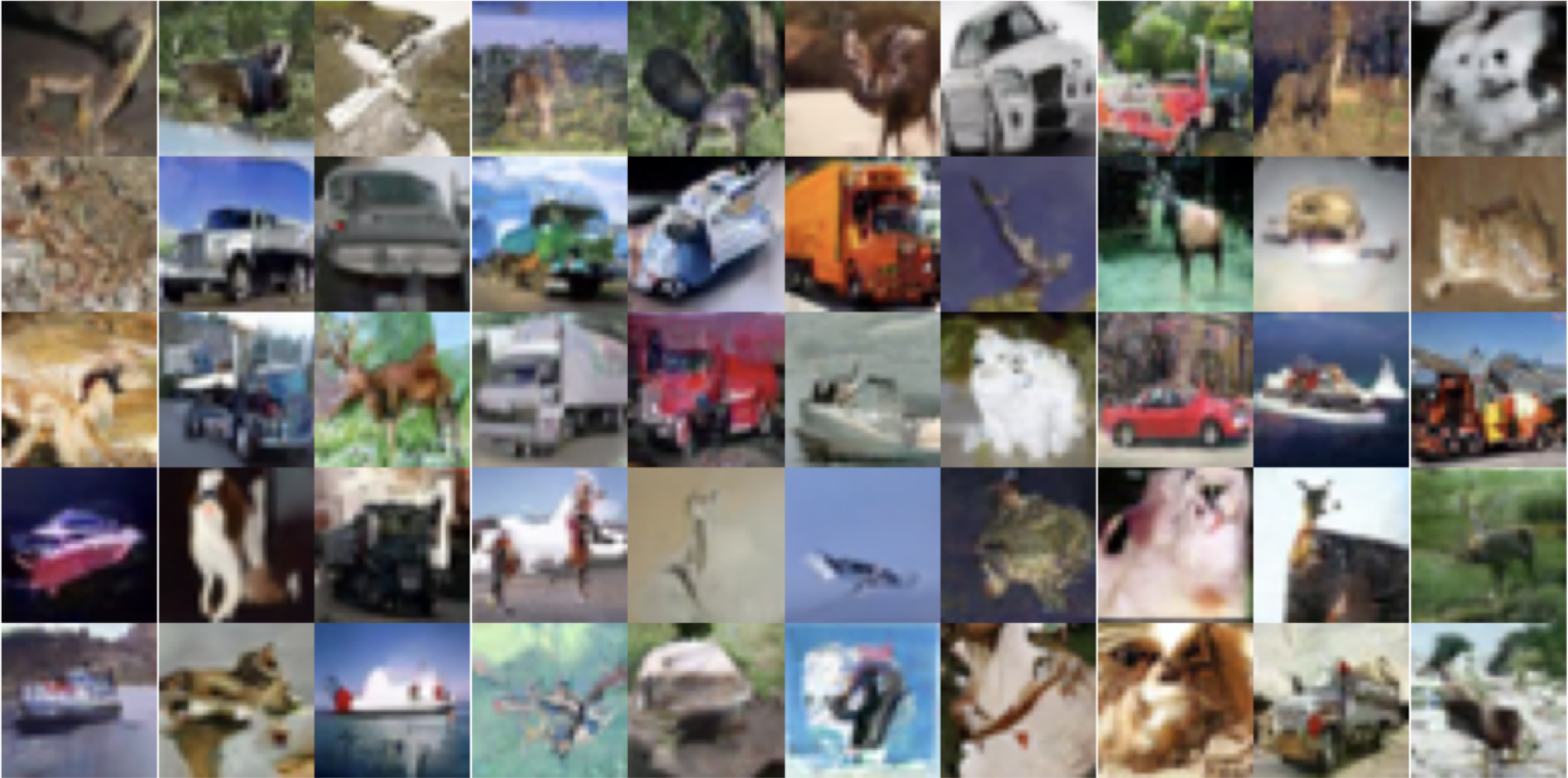}
    }
    \subfloat[Denoised (FID: 4.80)]{
        \includegraphics[width=0.48\textwidth]{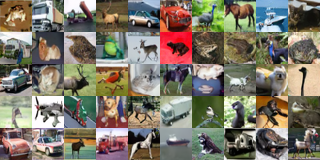}
    } 
    \caption{2,000 clean samples (4\%), noise level $\sigma = 0.59$.}
\end{figure}

\subsection{CelebA (fixed $\gamma$)}

\begin{figure}[H]
    \centering
    \subfloat[Generated (FID: 3.23)]{
        \includegraphics[width=0.48\textwidth]{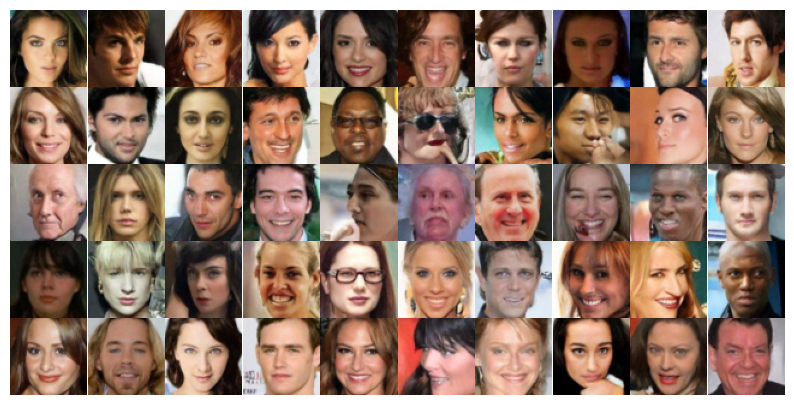}
    }
    \subfloat[Denoised (FID: 1.07)]{
        \includegraphics[width=0.48\textwidth]{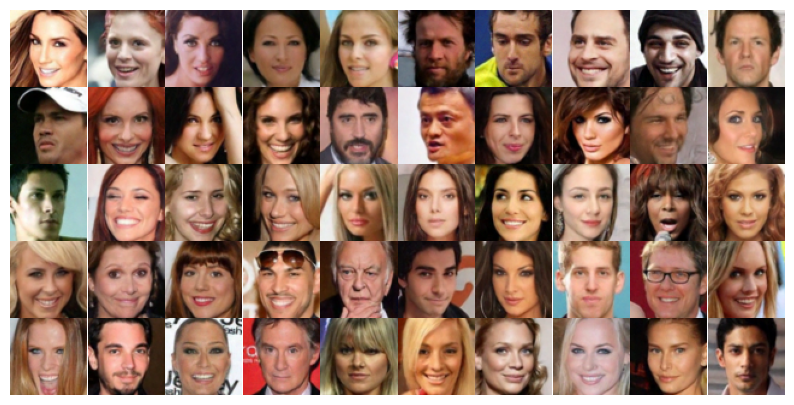}
    } 
    \caption{50 clean samples, noise level $\sigma = 0.2$.}
\end{figure}

\begin{figure}[H]
    \centering
    \subfloat[Generated (FID: 27.09)]{
        \includegraphics[width=0.48\textwidth]{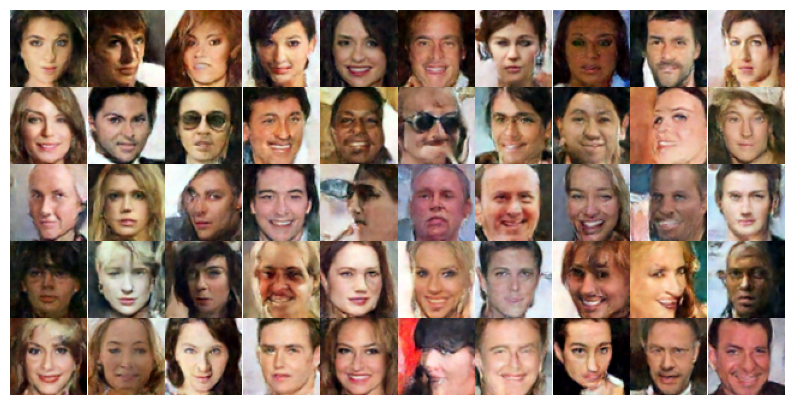}
    }
    \subfloat[Denoised (FID: 24.31)]{
        \includegraphics[width=0.48\textwidth]{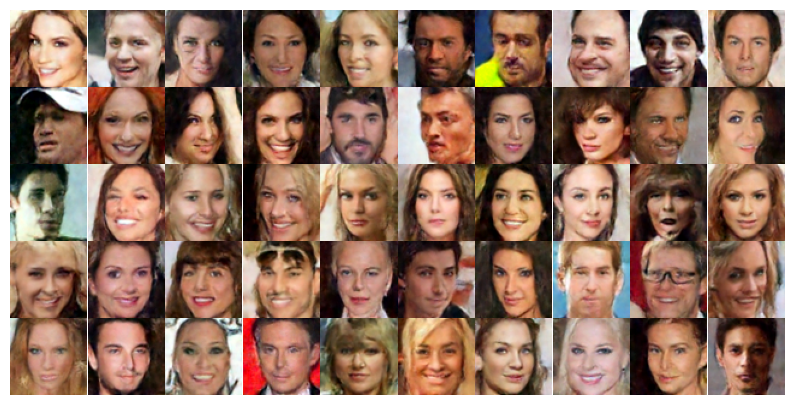}
    } 
    \caption{50 clean samples, noise level $\sigma = 1.38$.}
\end{figure}

\begin{figure}[H]
    \centering
    \subfloat[Generated (FID: 5.72)]{
        \includegraphics[width=0.48\textwidth]{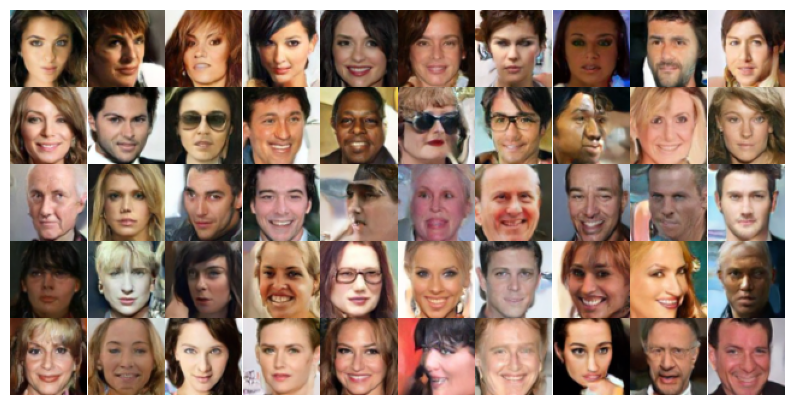}
    }
    \subfloat[Denoised (FID: 4.28)]{
        \includegraphics[width=0.48\textwidth]{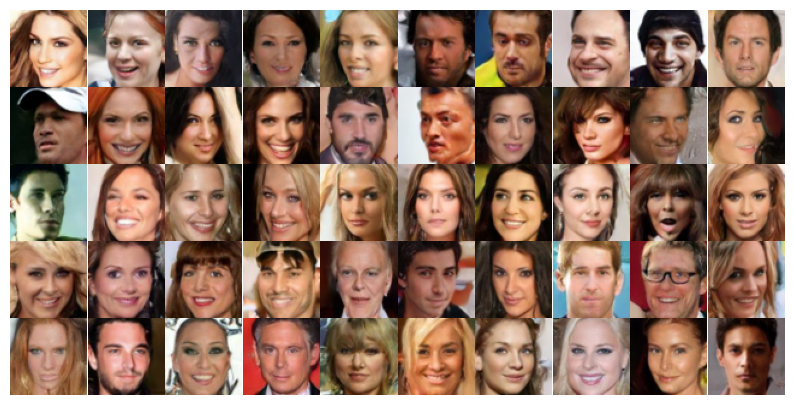}
    } 
    \caption{1,500 clean samples, noise level $\sigma = 1.38$.}
\end{figure}

\subsection{CelebA (adpative $\gamma$)}

\begin{figure}[H]
    \centering
    \subfloat[Generated (FID: 3.19)]{
        \includegraphics[width=0.48\textwidth]{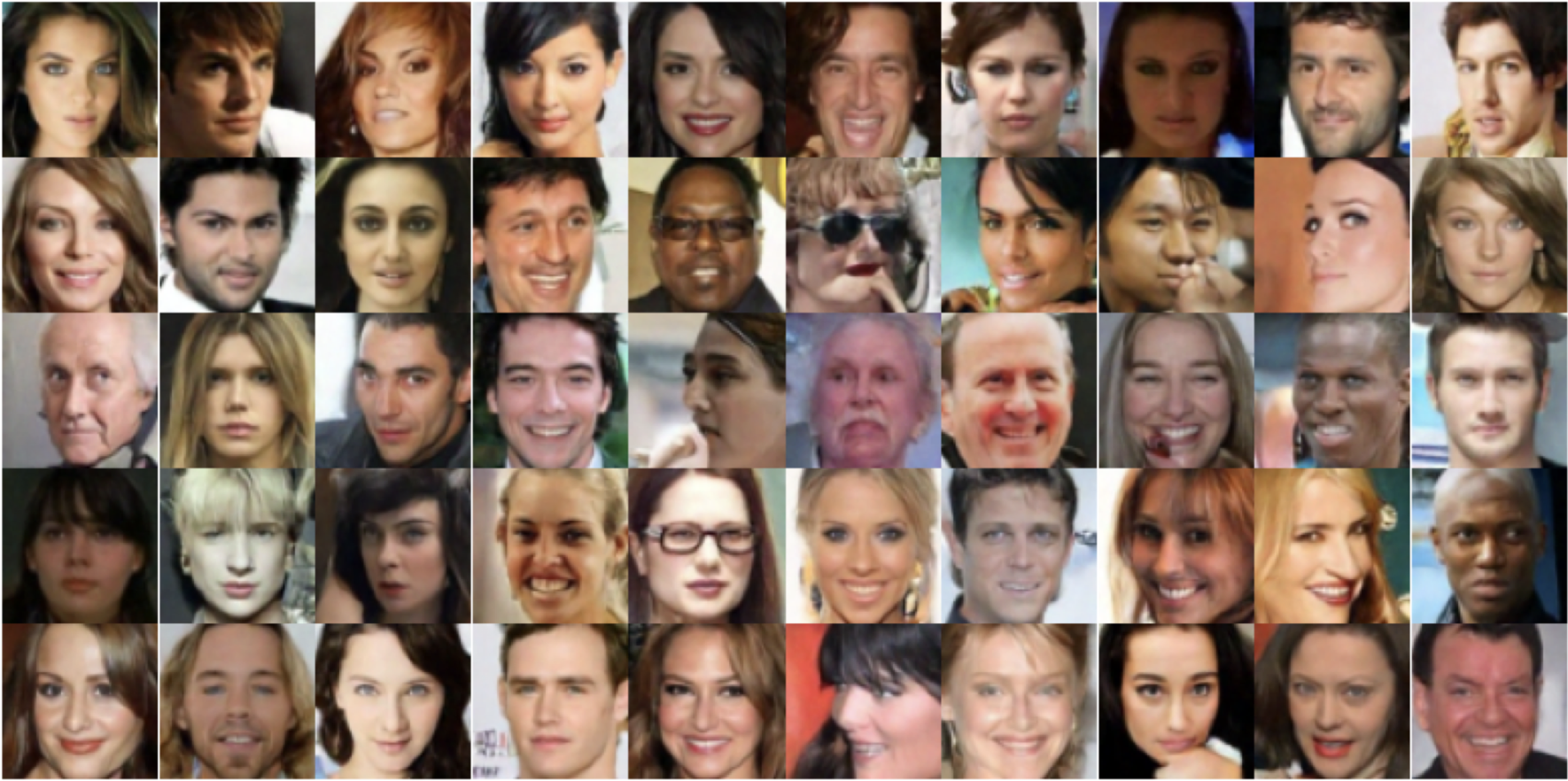}
    }
    \subfloat[Denoised (FID: 1.06)]{
        \includegraphics[width=0.48\textwidth]{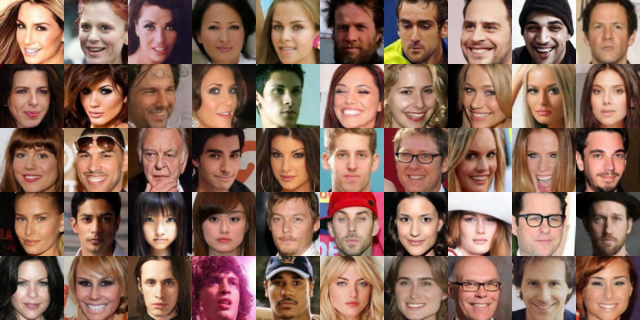}
    } 
    \caption{50 clean samples, noise level $\sigma = 0.2$.}
\end{figure}

\begin{figure}[H]
    \centering
    \subfloat[Generated (FID: 20.21)]{
        \includegraphics[width=0.48\textwidth]{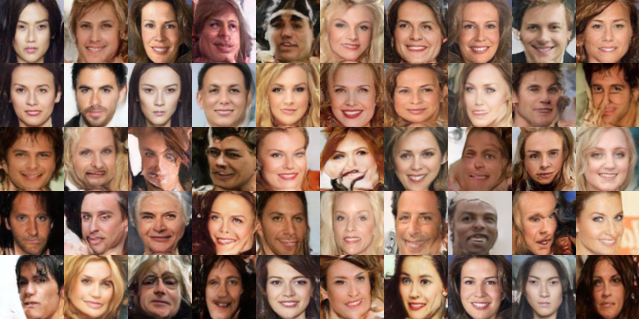}
    }
    \subfloat[Denoised (FID: 19.09)]{
        \includegraphics[width=0.48\textwidth]{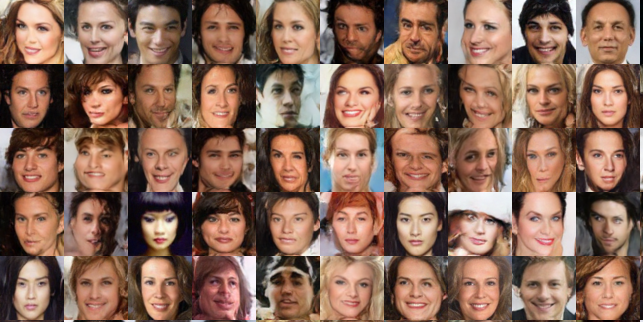}
    } 
    \caption{50 clean samples, noise level $\sigma = 1.38$.}
\end{figure}

\begin{figure}[H]
    \centering
    \subfloat[Generated (FID: 5.40)]{
        \includegraphics[width=0.48\textwidth]{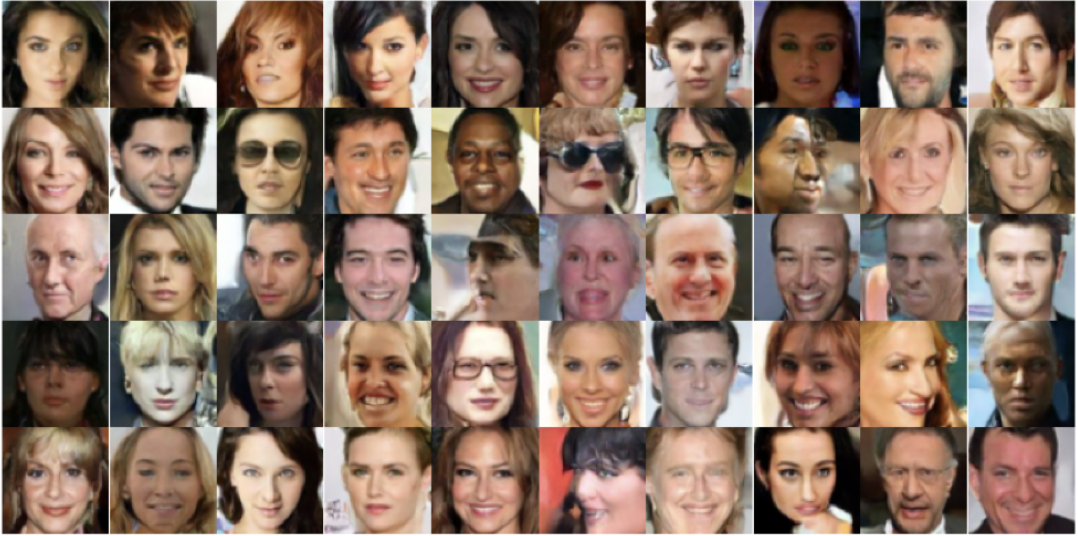}
    }
    \subfloat[Denoised (FID: 4.21)]{
        \includegraphics[width=0.48\textwidth]{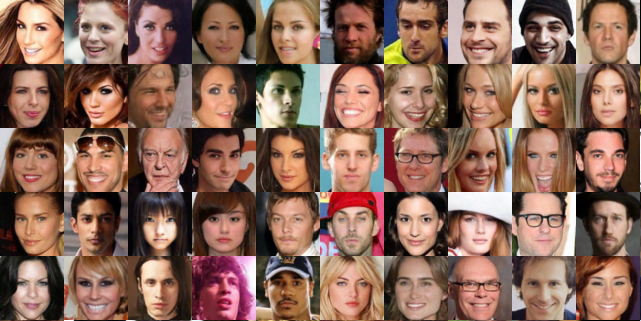}
    } 
    \caption{1,500 clean samples, noise level $\sigma = 1.38$.}
\end{figure}

\section{EXPERIMENT CONFIGURATIONS}
\label{appx:expConfig}
\subsection{Hardware Configurations}
\label{appx:hardware_config}

All diffusion models were trained on the main process using four NVIDIA A40 or RTX 6000 GPUs, managed by a SLURM scheduling system. The asynchronous denoising process ran concurrently in the background on a separate RTX 6000 GPU, taking less than 2.5 minutes to update 640 images on CIFAR-10 and under 5 minutes on CelebA.

Training on CIFAR-10 completes in under 5 days, and CelebA experiments in under 8 days.

\subsection{Model Architectures}
\label{appx:model_arch}
We implement the proposed Online SFBD algorithm using the EDM backbone~\citep{KarrasAAL22}, following the configuration described below throughout our empirical studies.
\begin{table}[H]
    \centering
    \caption{Experimental Configuration for CIFAR-10 and CelebA}
    \label{tab:experiment-config}
    \renewcommand{\arraystretch}{1.2} 
    \setlength{\tabcolsep}{6pt} 
    \resizebox{\textwidth}{!}{
    \begin{tabular}{p{4cm} p{5cm} p{5cm}} 
        \toprule
        \textbf{Parameter} & \textbf{CIFAR-10} & \textbf{CelebA} \\
        \midrule
        \textbf{General} & & \\
        Batch Size & 512 & 256 \\
        Loss Function & \texttt{EDMLoss} \citep{KarrasAAL22} & \texttt{EDMLoss} \citep{KarrasAAL22} \\
        Denoising Method &  $2^\text{nd}$ order Heun method (EDM) \citep{KarrasAAL22} & $2^\text{nd}$ order Heun method (EDM) \citep{KarrasAAL22}  \\
        Sampling Method &  $2^\text{nd}$ order Heun method (EDM) \citep{KarrasAAL22} & $2^\text{nd}$ order Heun method (EDM) \citep{KarrasAAL22}  \\
        Sampling steps & 18 & 40 \\
        \midrule 
        \textbf{Network Configuration} & & \\
        Dropout & 0.13 & 0.05 \\
        Channel Multipliers & $\{2, 2, 2\}$ & $\{1, 2, 2, 2\}$ \\
        Model Channels & 128 & 128 \\
        Resample Filter & $\{1, 1\}$ & $\{1, 3, 3, 1\}$ \\
        Channel Mult Noise & 1 & 2 \\
        \midrule
        \textbf{Optimizer Configuration} & & \\
        Optimizer Class & \texttt{RAdam} \citep{KingmaBa2014, LiuJHCLGH2020} & \texttt{RAdam}  \citep{KingmaBa2014, LiuJHCLGH2020} \\
        Learning Rate & 0.001 & 0.0002 \\
        Epsilon & $1 \times 10^{-8}$ & $1 \times 10^{-8}$ \\
        Betas & (0.9, 0.999) & (0.9, 0.999) \\
        \bottomrule
  \end{tabular}
  }%
\end{table}

\subsection{Datasets}
All experiments on CIFAR-10~\citep{Krizhevsky2009} and CelebA~\citep{LiuLWT2015} are conducted using only the training set. For FID evaluation, the model generates 50,000 samples, and FID is computed against the full training set, which includes both copyright-free and sensitive samples.



\end{document}